%% file: main.tex
\documentclass[10pt,twocolumn,letterpaper]{article}

\usepackage{cvpr}              
\usepackage{subcaption} 
\usepackage{placeins}
\usepackage{amsmath}
\usepackage{graphicx}
\usepackage{multicol,multirow}
\usepackage{float}

\usepackage{enumitem}
\newcommand{\comm}[2]{{\textcolor{#1}{#2}}}
\newcommand{\sg}[1]{\comm{black}{#1}}

\newcommand{\gm}[1]{\comm{black}{#1}}

\usepackage[normalem]{ulem}
\usepackage{siunitx}
\usepackage{xspace}
\makeatletter
\DeclareRobustCommand\onedot{\futurelet\@let@token\@onedot}
\def\@onedot{\ifx\@let@token.\else.\null\fi\xspace}
\DeclareRobustCommand\nodot{\futurelet\@let@token\@nodot}
\def\@nodot{\ifx\@let@token.\else~\null\fi\xspace}

\def\etal{\emph{et al}\onedot}
\makeatother

\newcommand{\name}{\textsf{LapisGS}}

\input{preamble}

\definecolor{cvprblue}{rgb}{0.21,0.49,0.74}
\usepackage[pagebackref,breaklinks,colorlinks,citecolor=cvprblue]{hyperref}

\def\paperID{137} 
\def\confName{3DV\xspace}
\def\confYear{2025\xspace}

\title{\raisebox{-0.1cm}{\includegraphics[scale=0.025]{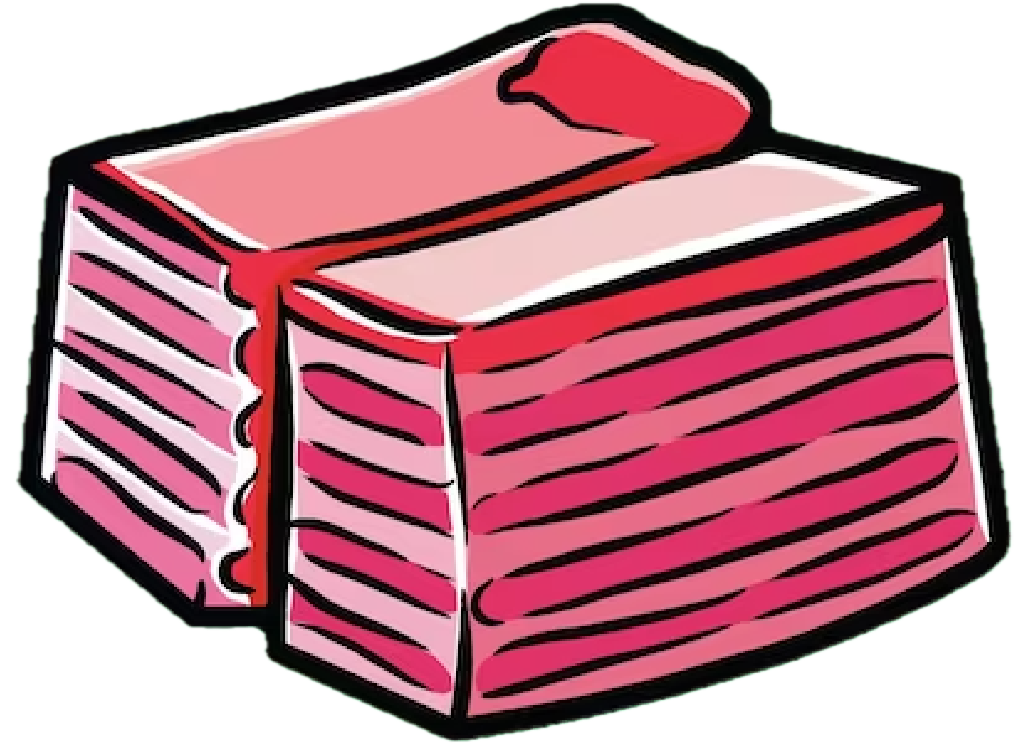}} {\name}: Layered Progressive 3D Gaussian Splatting\\ for Adaptive Streaming}

\author{%
  \hspace{0mm} Yuang Shi$^{1,3}$ \and Géraldine Morin$^{2,3}$ \and Simone Gasparini$^{2,3}$  \hspace{0mm} \and
   Wei Tsang Ooi$^{1,3}$
  \vspace{1mm}
  \and
  $^1$National University of Singapore \quad $^2$IRIT - University of Toulouse  \quad $^3$IPAL, IRL2955, Singapore \vspace{-1mm} \\ \vspace{-3mm}
  \makebox[0cm]{\tt\footnotesize \{yuangshi, ooiwt\}@comp.nus.edu.sg \{geraldine.morin, simone.gasparini\}@toulouse-inp.fr} \vspace{-2mm}
}

\begin{document}
\maketitle
\input{sec/0_abstract}    
\input{sec/1_intro}
\input{sec/2_related}

\input{sec/4_methodology}

\input{sec/5_experiments}

\input{sec/6_conclusion}

{
    \small
    \bibliographystyle{ieeenat_fullname}
    \bibliography{main}
}

\input{sec/X_suppl}
\end{document}

%% file: preamble.tex
\usepackage[dvipsnames]{xcolor}


%% file: sec/0_abstract.tex
\begin{abstract}
The rise of Extended Reality (XR) requires efficient streaming of 3D online worlds, challenging current 3DGS representations to adapt to bandwidth-constrained environments. 
This paper proposes {\name}, a layered 3DGS that supports adaptive streaming and progressive rendering.
Our method constructs a layered structure for cumulative representation, incorporates dynamic opacity optimization to maintain visual fidelity, and utilizes occupancy maps to efficiently manage Gaussian splats. 
This proposed model offers a progressive representation supporting a continuous rendering quality adapted for bandwidth-aware streaming.
Extensive experiments validate the effectiveness of our approach in balancing visual fidelity with the compactness of the model, with up to 50.71\% improvement in SSIM, 286.53\% improvement in LPIPS with 23\% of the original model size, and shows its potential for bandwidth-adapted 3D streaming and rendering applications. Project page: \url{https://yuang-ian.github.io/lapisgs/}
\end{abstract}

%% file: sec/1_intro.tex
\section{Introduction}
\label{sec:intro}

In recent years, there has been a significant surge in the popularity of diverse XR applications like virtual reality (VR), augmented reality (AR), and cloud gaming, largely driven by the need to provide users with seamless access to online 3D environments.
The success of these applications hinges on the ability to represent complex 3D scenes accurately.

Recently, 3D Gaussian Splatting (3DGS) \cite{kerbl20233d} has emerged as an efficient technique for generating an explicit 3D representation from calibrated images, providing photo-realistic visual quality and supporting fast rendering.

\begin{figure}
    \centering
    \includegraphics[width=\linewidth]{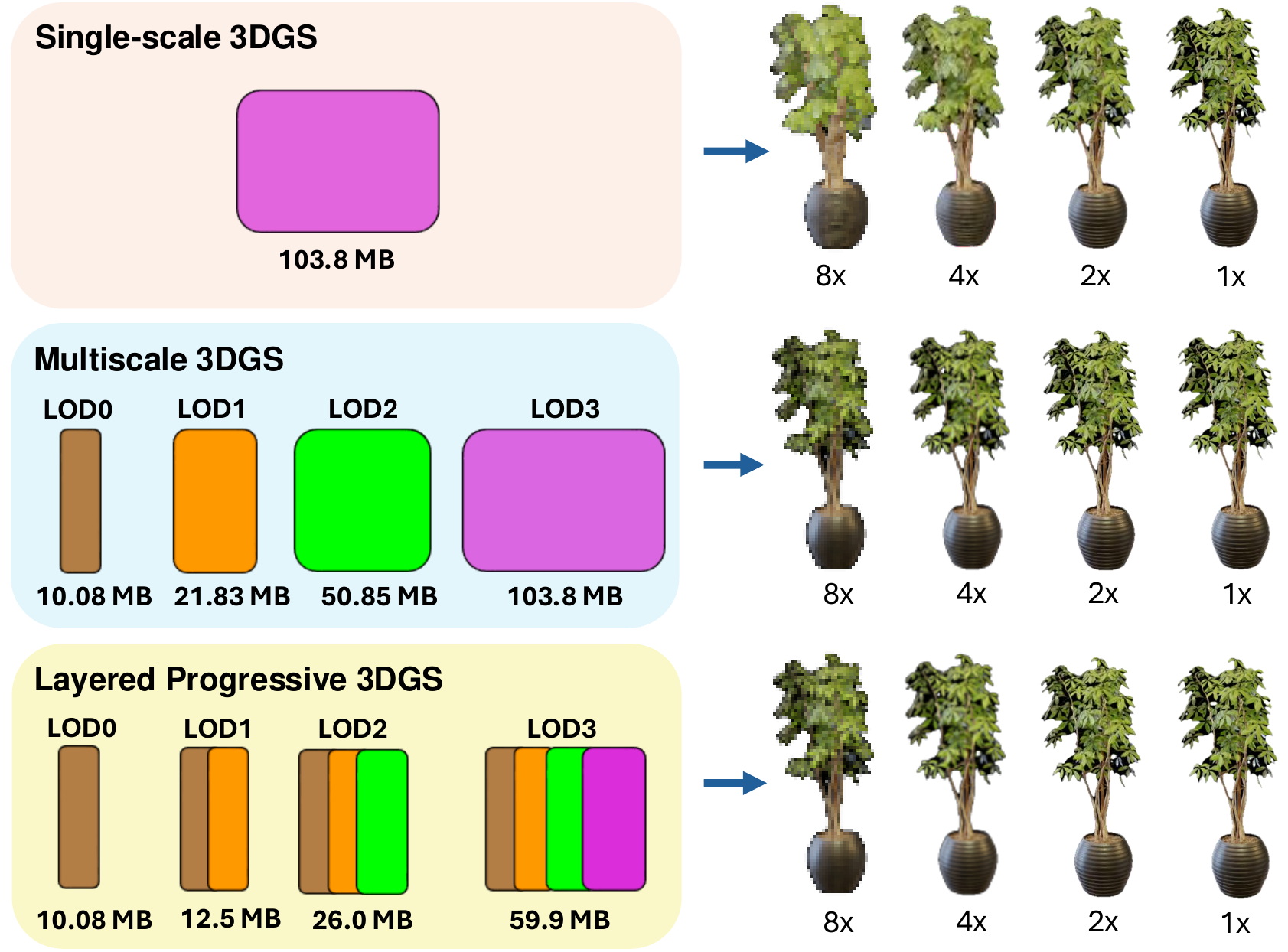}
    \caption{The illustration of the architecture of the single-scale model (upper), multiscale model (middle), and layered progressive model (lower) and their sample renderings at different resolution scales (1$\times$, 2$\times$, 4$\times$, and 8$\times$). Our layered progressive model is tailored for adaptive streaming and view-adaptive rendering.}
    \label{fig:teaser}
\end{figure}

While 3DGS offers advantages for representing and rendering high-fidelity 3D scenes, the resulting data size can be prohibitively large for streaming to users with various bandwidth constraints due to transfer over diverse and dynamic networks and different rendering capacities due to heterogeneous devices.
Recent works~\cite{fan2023lightgaussian,hac2024,morgenstern2023compact,navaneet2023compact3d,papantonakis24reduce,lee2024compact,lu2024scaffold,shi2025sketch} to reduce data size through model compaction or optimization are insufficient to address the dynamic nature of network conditions and device capabilities.

To overcome these limitations, adaptive streaming is crucial to optimize resource utilization, visual quality, and user experience~\cite{shi2024qv4}.
Specifically, adaptive streaming enables efficient and scalable transmission of 3D content by leveraging a layered representation.
This layered structure organizes the 3D data into multiple levels of detail (LOD), typically comprising a base layer and one or more enhancement layers.
The base layer provides a basic representation, while the enhancement layers progressively refine the visual quality.

The layered representation must meet specific requirements to effectively support adaptive streaming for 3DGS.
Firstly, given subsets of Gaussian splats, the model should be able to represent complete 3D content with reduced details, allowing for the generation of a lower-quality base and enhancement layers.
Secondly, progressive levels of detail should inherently share visual information 
with lower levels to minimize redundancy in streaming and support view-adaptive rendering.
By leveraging shared visual characteristics across different levels, the progressive representation can reduce the amount of data needed to be transmitted while facilitating smooth and continuous level transition for rendering.

\begin{figure}[t]%
    \centering
    \begin{subfigure}{0.24\linewidth}
        \includegraphics[width=\linewidth]{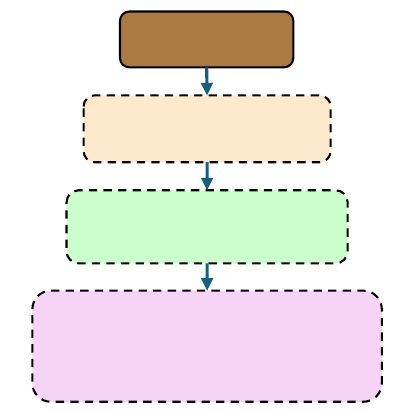}
        \caption{LOD 0.}
    \end{subfigure}
    \begin{subfigure}{0.24\linewidth}
        \includegraphics[width=\linewidth]{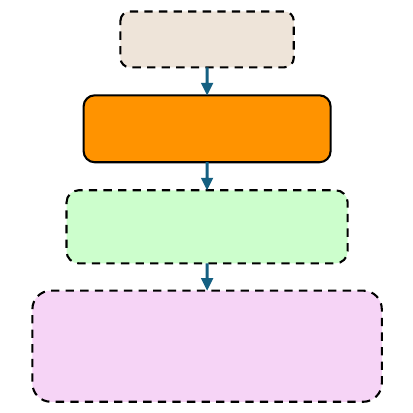}
        \caption{LOD 1.}
    \end{subfigure}
    \begin{subfigure}{0.24\linewidth}
        \includegraphics[width=\linewidth]{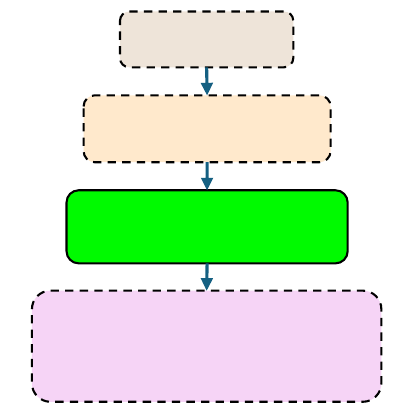}
        \caption{LOD 2.}
    \end{subfigure}
    \begin{subfigure}{0.24\linewidth}
        \includegraphics[width=\linewidth]{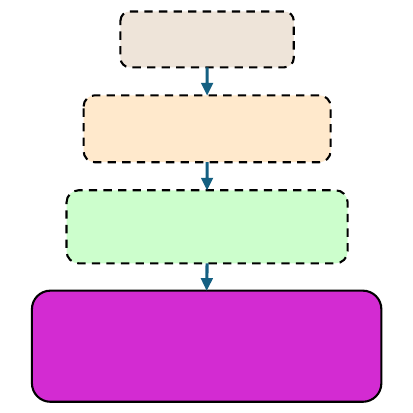}
        \caption{LOD 3.}
    \end{subfigure} 
    \\ 
    \vspace{0.5cm} 
    \begin{subfigure}{0.24\linewidth}
        \includegraphics[width=\linewidth]{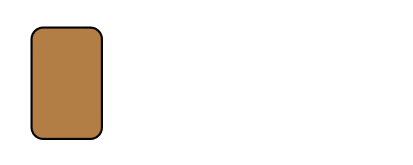}
        \caption{LOD 0.}
    \end{subfigure}
    \begin{subfigure}{0.24\linewidth}
        \includegraphics[width=\linewidth]{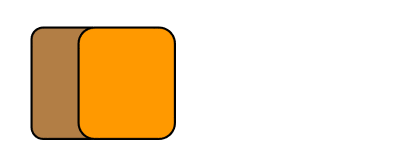}
        \caption{LOD 1.}
    \end{subfigure}
    \begin{subfigure}{0.24\linewidth}
        \includegraphics[width=\linewidth]{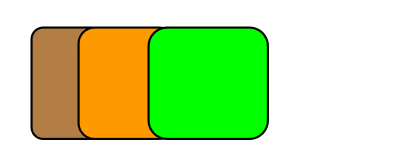}
        \caption{LOD 2.}
    \end{subfigure}
    \begin{subfigure}{0.24\linewidth}
        \includegraphics[width=\linewidth]{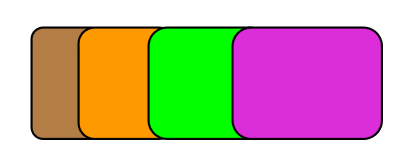}
        \caption{LOD 3.}
    \end{subfigure} 
    \caption{Illustrations of space-based representation characterized by discrete quality layers (upper) and layered progressive representation which comprises incremental layers (lower).}%
    \label{fig:hierarchy_illus}
\end{figure}

Recent efforts have been made to build hierarchical 3DGS representations based on multiresolution space partitioning to render very large scenes in real time~\cite{kerbl2024hierarchical,ren2024octree,liu2024citygaussian}. 
For instance, Ren \etal~\cite{ren2024octree} decompose the 3D space with an octree and optimize Gaussian splats at each octree node, while Kerbl \etal~\cite{kerbl2024hierarchical} regard every splat as a tree node and construct a tree-based hierarchy by recursively merging the neighboring Gaussian splats into parent nodes.
However, these space-based hierarchies, while being effective and efficient for local rendering, suffer from the following drawbacks in supporting adaptive streaming:
\begin{itemize}
    \item Being built based on spatial information, space-based hierarchies do not explicitly define quality levels. Hence, tree traversal algorithms are required to select appropriate splats at different layers to build levels of detail.
    This added complexity can be computationally expensive and less adaptable to rapidly changing network conditions, potentially leading to inefficient streaming decisions. 
    \item Secondly, each level of detail captured from the space-based hierarchy is independently represented by a set of anchor splats, as shown in \cref{fig:hierarchy_illus}.
    The lack of correlation between levels can hinder efficient progressive transmission and limit flexible detail adjustment across different parts of the scene, such as foveated rendering or distance-aware rendering~\cite{shi2024perceptual, li2022progressive,schutz2019real}.
    \item Furthermore, since different levels of detail do not share content, each level needs to redundantly encode and represent similar visual information using separate sets of Gaussian splats, significantly enlarging both the global and intermediate \gm{model size}, as shown in Figure \ref{fig:teaser}.
\end{itemize}

In this paper, we propose {\name}\footnote{\textit{Lapis} means "layer" in Malay, the national language of Singapore --- the host of 3DV'25.   The logo in the title depicts \textit{kuih lapis}, or "layered cake", a local delight in Singapore and neighboring countries}, a layered representation for progressive streaming and rendering of 3DGS content.
Inspired by scalable coding~\cite{schwarz2007overview,wang2021multiscale,ahn2012efficient} and progressive LOD representation~\cite{li2022progressive,xiangli2022bungeenerf,zhu2023pyramid,schwarz2007overview,schutz2019real}, this method is designed to efficiently stream and render photo-realistic 3D objects and scenes by leveraging a progressive framework that enables dynamic adaptation to varying levels of detail.
As shown in \cref{fig:hierarchy_illus}, at the core of our approach is a layered structure for cumulative representation, where each layer adds additional details to the existing base layers, progressively refining the representation. 
To force coherence among layers while avoiding re-encoding lower layers, we incorporate dynamic opacity optimization during training, allowing for selective adjustment of layer contributions for optimal visual fidelity. 
We then utilize an occupancy map to track and exclude less important Gaussian splats during streaming and rendering, improving computational and storage efficiency. 

Our contributions can be summarized as follows.

\begin{itemize}
    \item A progressive layered approach for 3DGS encoding multiple levels of detail into a single-layered model, supporting adaptive streaming and seamless rendering.
    \item Dynamic opacity optimization and management, which ensures consistency across varying resolution levels but also adjusts layer contributions selectively to maintain visual fidelity.
    As a result, our approach reduces data size and additionally enhances computational efficiency by managing splats dynamically with occupancy maps.
    \item Flexible and adaptive rendering, which enables seamless transitions and view-adaptive rendering strategies without the need for separate models for each level of detail.
    \item Extensive experiments on diverse 3D contents demonstrate the effectiveness of our method, achieving high-quality rendering and low resource cost, with up to 50.71\% improvement in SSIM, 286.53\% improvement in LPIPS with 23\% of the original model size.
\end{itemize}

By drawing parallels to scalable coding and LOD representation, our progressive representation effectively balances the need for high-quality rendering with bandwidth-aware streaming, providing a robust solution for adaptive 3D content delivery. 

To foster collaboration and further research, we will release our source code to the research community and make our pre-trained layered 3DGS models publicly available.

%% file: sec/2_related.tex
\section{Background and Related Work} \label{sec:related}

\subsection{3D Gaussian Splatting} \label{subsec:3dgs}

3D Gaussian Splatting (3DGS)~\cite{kerbl20233d} is a method used for real-time photorealistic radiance field rendering by representing a 3D scene as a collection of 3D Gaussian splats.
Each 3D Gaussian splat is characterized by a set of attributes, including its position $\mathbf{x}$, a covariance matrix $\mathbf{\Sigma}$ decomposed by a scaling matrix $\mathbf{S}$ and a rotation matrix $\mathbf{R}$, opacity $\sigma$, and view-dependent color $c$ represented by a set of Spherical Harmonic (SH) coefficients.
Specifically, each 3D Gaussian is defined as $G(\mathbf{x})=exp(-\frac{1}{2}(\mathbf{x})^T \mathbf{\Sigma}^{-1}(\mathbf{x})).$ 
In the rendering stage~\cite{zwicker2001ewa}, each 3D Gaussian is projected into 2D camera coordinates, denoted as $G^{\prime}(\mathbf{x}^{\prime})$. To compute the color of a pixel $\mathbf{x}^{\prime}$, a tile-based rasterizer is employed to sort the projected splats in front-to-back depth order and blend their colors with $\alpha$-blending. The $\alpha$-blending weights are deprived as $\sigma G^{\prime}(\mathbf{x}^{\prime})$.

These Gaussian splats are derived from a sparse Structure-from-Motion (SfM) point cloud and are refined through a gradient-descent-based optimization process interleaved with adaptive refinements.



\subsection{Level-of-Detail 3DGS} \label{subsec:lod_3dgs}

Level-of-details (LODs) are fundamental to scalable rendering solutions, allowing for the adjustment of detail levels based on computational resources and user needs. While LODs have been explored for point cloud representations, applying them to 3DGS presents unique challenges.

Recent efforts have been made to build multiscale representations for 3DGS~\cite{kerbl2024hierarchical,ren2024octree,yan2024multi,liu2024citygaussian,chen2024gaussianeditor,lu2024scaffold}. 
For instance, Yan \etal~\cite{yan2024multi} proposed to add larger and coarser Gaussian splats for lower resolutions by aggregating the smaller and finer Gaussians from higher resolutions, creating independent Gaussian splats layers at different scales to mitigate the aliasing artifacts during rendering.
To reconstruct and render large-scale scenes with 3DGS, Liu \etal~\cite{liu2024citygaussian} divided the whole scene into spatially adjacent blocks and trained each block in parallel.
Within each block, different LODs were created with LightGaussian~\cite{fan2023lightgaussian}, by pruning and post-optimizing the splats.
Kerbl \etal \cite{kerbl2024hierarchical} proposed a tree-based hierarchy designed for real-time rendering of large-scale scenes. 
Their approach involves dividing the scene into chunks and constructing a tree for each chunk, where both interior and leaf nodes are represented by Gaussian splats. Interior nodes are formed by merging child splats and then optimized at each level separately, while leaf nodes originate directly from the initial optimization process.
To ensure smooth transitions between levels, splat interpolation is employed. 
Similarly, Ren \etal \cite{ren2024octree} use an octree structure to partition the 3D space, with each octree level corresponding to a set of anchor Gaussians that define the LOD. They incorporate Scaffold-GS~\cite{lu2024scaffold}, which leverages neural Gaussians and MLPs to predict anchor-wise features, allowing for a more compact representation. Linear interpolation of rendered 2D images is used to achieve smooth transitions between levels.  

However, these existing methods are primarily designed for reconstruction quality and rendering speed, without considering the challenges of adaptive streaming, as we discussed in \cref{sec:intro}. Space-based hierarchies in 3DGS feature discrete quality layers, each independently represented by a set of anchor splats, as shown in \cref{fig:hierarchy_illus}. This approach requires substantial computational resources for maintenance and navigation, especially with frequent level transitions, and limits view-adaptive rendering. In contrast, our model supports progressive streaming and flexible detail adjustment, reducing storage needs while enabling smoother level transitions and effective view-adaptive rendering.


%% file: sec/4_methodology.tex
\begin{figure}[t]
    \centering
    \includegraphics[width=0.95\linewidth]{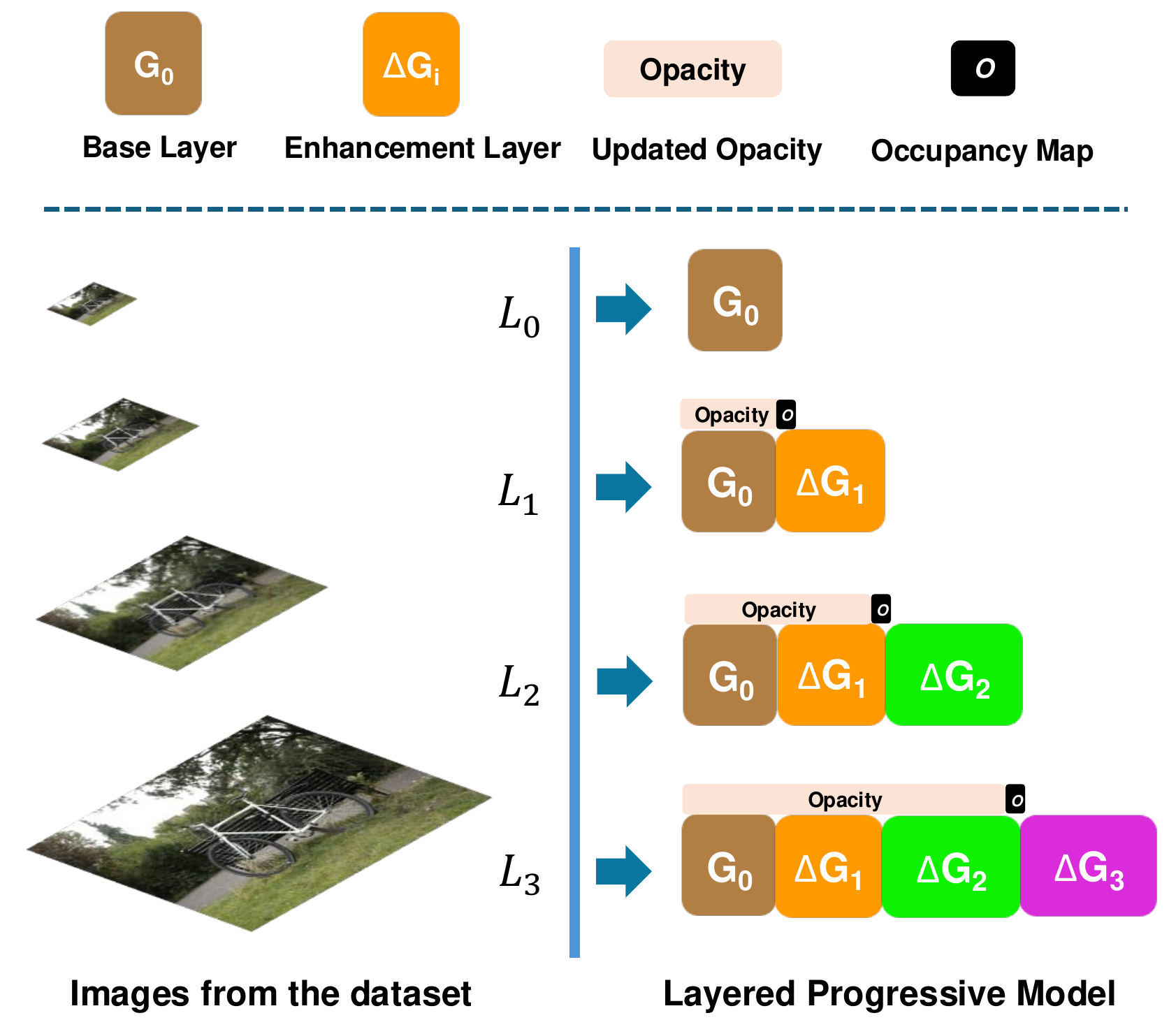}
    \caption{The overview of {\name}. The framework progressively constructs a layered model, starting from the low-resolution base layer ($L_0$) and adding higher-resolution enhancement layers ($L_1$, $L_2$, $L_3$). Dynamic opacity optimization and occupancy maps are employed to refine splat contributions and optimize data size.}
    \label{fig:methodology}
\end{figure}


\section{Methodology}
\label{sec:methodology}


Our method {\name} is based on training a 3DGS model at successive level of {progressively higher} resolution to create a multiscale representation. Initially, a low-resolution dataset is used to establish a base layer. As the training progresses, new enhancement layers are added, each trained on incrementally higher resolution versions of the dataset. These layers build upon and refine the details captured in the previous layers. 
While the parameters of the prior layers remain fixed, their opacity is optimized to adjust the influence of each layer dynamically.
We also utilize occupancy maps to track the contributions of splats.
During streaming and rendering, these transparent splats are excluded, which reduces the overall model size and improves computational efficiency. 
To ensure smooth transitions between resolution levels, we employ interpolation of opacity values between adjacent layers. 
\cref{fig:methodology} shows the overview of {\name}.

\subsection{Layered Progressive 3DGS} \label{subsec:progressive_3dgs}

\textbf{Layered Progressive Representation}.
We denote $N+1$ as the total number of quality levels for the layered 3DGS model, which also corresponds to the total number of training stages in the progressive training scheme.
We can then denote $L_{i}$ as the $i$-th level of detail where $i \in \{0,1,\ldots, N\}$. 
Given a full-resolution multi-view image set $\mathbf{D}_{N}$, we can build an image pyramid $\{\mathbf{D}_{i}\}^{N}_{i=0}$:
\begin{equation}
    \mathbf{D}_i = \left\{\left(V_m^i, X_m^i\right)\right\}_{m=1}^M, 
\end{equation}
where $V_m^i$ is the camera matrix, $X_m^i$ is the corresponding image, and $M$ is the number of views.

Starting from the lowest quality level, we initially optimize a set of 3D Gaussian splats, denoted as $\mathbf{G}_0$.
As training advances, views with higher resolution are considered at each layer.
Consequently, the Gaussian splats at various quality levels are represented as $\{\mathbf{G}_{i}\}^{N}_{i=0}$.

As discussed in \cref{sec:intro}, the core idea is to create a layered structure comprising a base layer $G_0$ and enhancement layers.
As training progresses to a higher quality level, new enhancement layers are optimized and integrated with prior layers. 
Formally, we can represent the layered progressive model as $\{\mathbf{G}_{0}, \{\Delta\mathbf{G}_{i}\}^{N}_{i=1}\}$, where 
\begin{equation}
    \mathbf{G}_i = \mathbf{G}_0 + \sum_{k=1}^{i} \Delta \mathbf{G_k}, i \in \{1,2,\ldots,N\},
\end{equation}
where $\Delta \mathbf{G_k}$ is the $k$-th enhancement layer.

\begin{figure}[t]%
    \centering
    \begin{subfigure}{0.24\linewidth}
        \includegraphics[width=\linewidth]{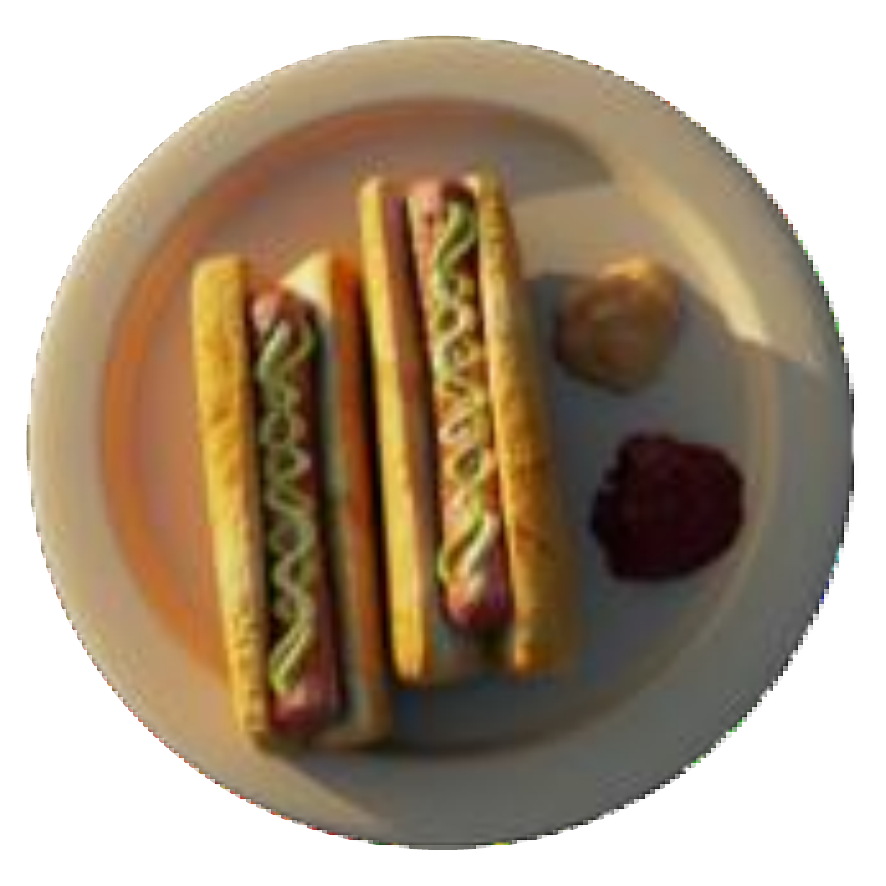}
        \caption{$\mathbf{G}_{1}$.}
    \end{subfigure}
    \begin{subfigure}{0.24\linewidth}
        \includegraphics[width=\linewidth]{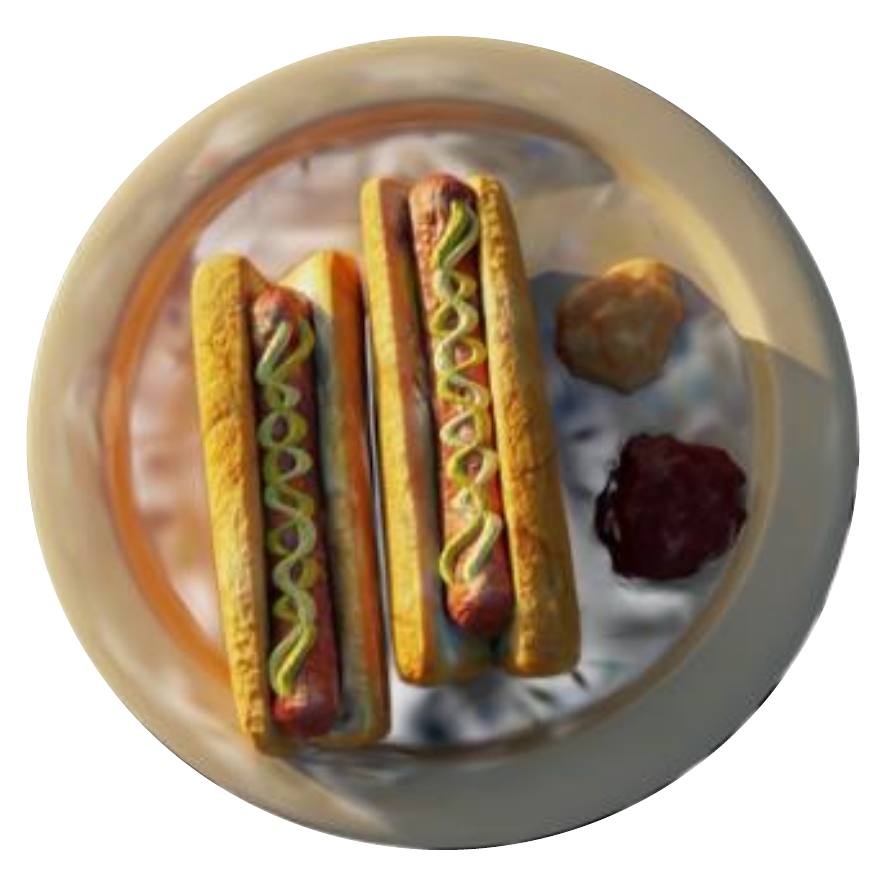}
        \caption{$\Delta \mathbf{G}_{2}$.}
    \end{subfigure}
    \begin{subfigure}{0.24\linewidth}
        \includegraphics[width=\linewidth]{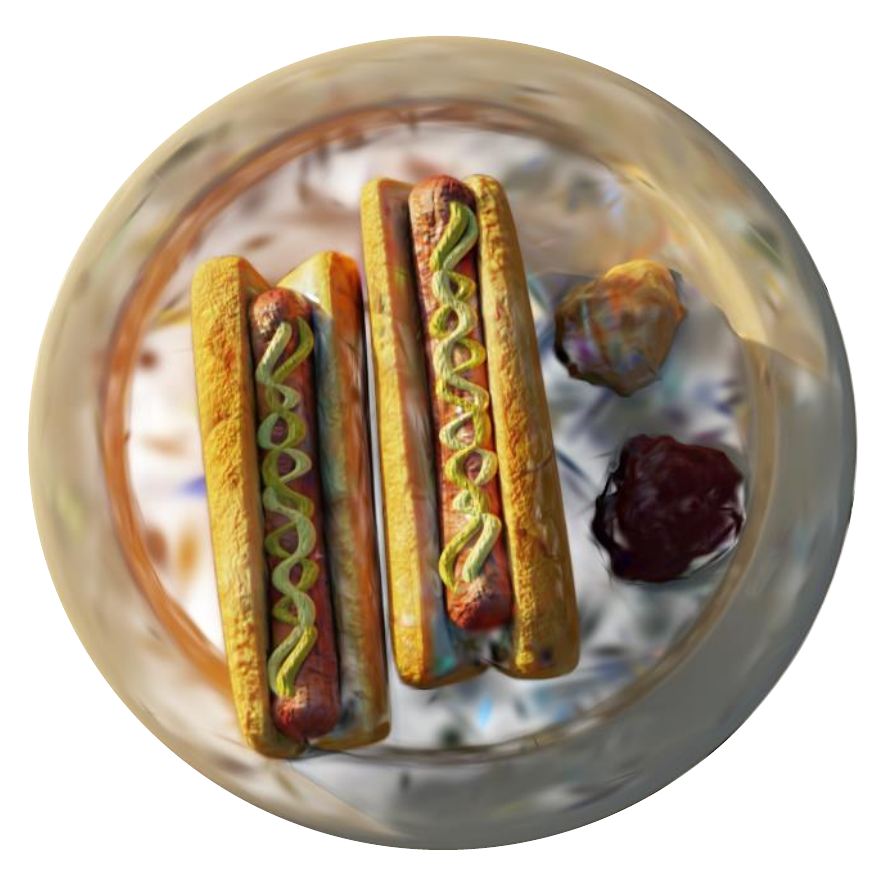}
        \caption{$\Delta \mathbf{G}_{3}$.}
    \end{subfigure}
    \begin{subfigure}{0.24\linewidth}
        \includegraphics[width=\linewidth]{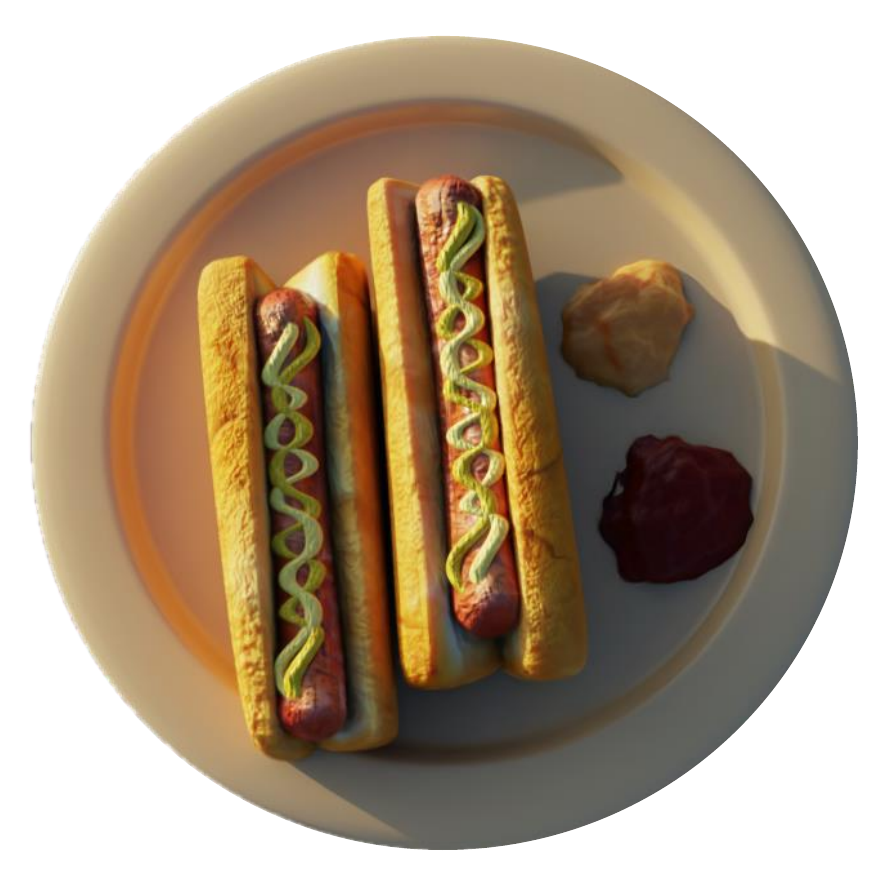}
        \caption{$\mathbf{G}_{3}$.}
    \end{subfigure}
    \caption{Sample renderings of \textit{hotdog}. The enhancement layers $\Delta \mathbf{G}_{2}$ and $\Delta \mathbf{G}_{3}$ capture higher frequency features and can be iteratively added to the layer $\mathbf{G}_{1}$ to obtain the model $\mathbf{G}_{3}$.}%
    \label{fig:progressive_illu}
\end{figure}

In this layered structure and progressive training scheme, a rough scene layout is constructed, allowing lower-frequency features to be learned in the early stages. 
This layout serves as a foundation for higher quality levels in subsequent training stages.
By building upon information from previous levels, the model can focus more on capturing higher frequency features, thereby speeding up convergence and reducing redundancy across different quality levels.
\cref{fig:progressive_illu} shows an example of the above process.

\textbf{Multi-level Optimization}.
During the training of each level in {\name}, we focus on optimizing the enhancement layer $\Delta\mathbf{G}_i$ while maintaining the parameters of the preceding layers $\{\mathbf{G}_0, \Delta\mathbf{G}_1, \ldots, \Delta\mathbf{G}_{i-1}\}$ as fixed, except for their opacity values.

The decision to optimize only the opacity values of the previous layers, rather than other parameters, is based on the balance between efficient layer integration and visual coherence. 
3DGS applies standard $\alpha$-blending for rendering, indicating that the contribution of each splat is typically additive and weighted by its opacity. 
Besides, unlike other attributes of Gaussian splats (such as position or scale), changing opacity does not alter the spatial information of the scene~\cite{fan2023lightgaussian,ren2024octree,chen2024text,sun2024multi}. 
Therefore, opacity optimization allows for fine-tuning the visibility and influence of Gaussian splats from earlier layers without altering their foundational features and changing the layered structure.
In other words, this selective optimization strategy enables the model to refine the contributions of existing layers, ensuring that previously learned features are kept, while the current enhancement layer focuses on capturing new details and improving the overall representation.

\begin{figure*}[t]%
    \centering
    \includegraphics[width=0.95\linewidth]{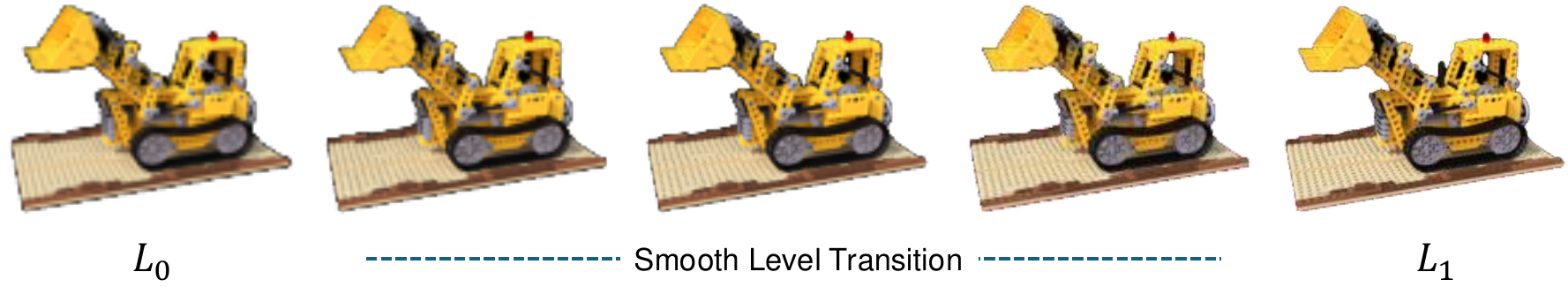}
    \caption[]{Example of the smooth level transition on \textit{Lego}. }%
    \label{fig:smooth_transition}
\end{figure*}

For each level $L_i$, the optimization process is driven by minimizing a rendering loss function $\mathcal{L}_{\Delta \mathbf{G}_i}$, which consists of two components: an $L_1$-norm loss $\mathcal{L}_1$ and a D-SSIM loss $\mathcal{L}_{\mathrm{D}}$.
The rendering loss is defined as follows:
\begin{equation}
    \mathcal{L}_{\Delta \mathbf{G}_{i}} = \lambda\sum_{m=1}^M \mathcal{L}_1(\hat{X}_m^i, X_m^i) + (1-\lambda)\sum_{m=1}^M \mathcal{L}_{\mathrm{D}}(\hat{X}_m^i, X_m^i),
    \label{eq:loss}
\end{equation}
where $X_m^i$ represents the ground truth image for view $m$ at the $i$-th quality level, and $\hat{X}_m^i$ is the image rendered of $\mathbf{G}_i$, using all enhancement layers up to $L_i$. 

Notably, $\mathcal{L}_1$ loss estimates perceived errors and is not sensitive to blurriness or low-resolution artifacts that do not alter the image's structure~\cite{wang2004image,zhang2024fregs}. In our progressive training pipeline, this insensitivity can result in premature convergence, where the model fails to update and densify the ``low-layer'' Gaussian splats. 
Existing works~\cite{zhang2024fregs,yu2024mip,chen2024frequency,yan2024multi} offer sophisticated solutions. For example, Zhang \etal~\cite{zhang2024fregs} propose to utilize frequency information to regularize the Gaussian densification, by incorporating frequency term in the loss function.
Nevertheless, addressing this problem is orthogonal to our work. 
In our approach, we adopt a straightforward step to alleviate the problem by giving more weight to the SSIM loss with $\lambda=0.2$, to ensure that the model prioritizes maintaining structural integrity.


By minimizing the rendering loss at each level, the model incrementally refines its representation, effectively balancing the integration of new information and the learned features of lower-resolution layers.
This approach ensures that the final model achieves high visual fidelity and efficient splat coding across successive quality levels.

\textbf{Representation Compaction and Adjustment}.
Gaussian splats from the first layers capture broader and low-frequency features essential for constructing a rough scene layout.
As the training process progresses to higher resolutions, the model focuses on capturing fine-grained details that the lower-resolution splats cannot effectively represent.
Consequently, the opacity of these ``low-layer'' splats is reduced during the optimization of ``high-layer'' splats, indicating their decreased importance in the rendering process.

To improve transmission and computational efficiency, we introduce an occupancy map $O_i$ for each level $L_i$ in our layered 3DGS model.
This map tracks the opacity of individual Gaussian splats, identifying those that fall below a specified opacity threshold.
These splats are marked as less important and can be omitted during both streaming and rendering processes, significantly reducing the model size during transmission and rendering.
We set the opacity threshold at $0.005$, following the default pruning phase setting from the original 3DGS work~\cite{kerbl20233d}.

Nevertheless, this approach could allow for adaptive bitrate streaming by dynamically selecting splats based on their opacity.
By adjusting the opacity threshold, the system could fine-tune the amount of data transmitted, adapting to varying network bandwidths and device capabilities~\cite{sun2024multi, fan2023lightgaussian}. 

\subsection{Continuous level transition}
A critical challenge in LOD rendering is achieving seamless visual transitions between different resolution levels.
Abrupt changes between levels can lead to visual artifacts, degrading the user experience, especially in scenarios with limited bandwidth or dynamic view changes~\cite{sun2024multi}. 

We address this by linearly interpolating the opacity of a given splat between adjacent levels.
As discussed in \cref{subsec:progressive_3dgs} and \cref{subsec:3dgs}, opacity leverages each splat's contribution during rendering, and by adjusting these weights, we effectively blend representations from different quality levels in a plausible manner.
Interpolating the opacity allows for gradual detail blending without altering spatial information, resulting in smooth visual changes. Our method aligns with the additive nature of Gaussian splat rendering and leverages the human visual system's lower sensitivity to gradual intensity changes.

Specifically, for a target resolution $r_t$ that falls between the resolutions of two adjacent quality levels $L_i$ and $L_{i+1}$, we define a continuous interpolation factor $t$.
This factor is calculated as 
\begin{equation}
    t(r_t) = \frac{r_t - r_i}{r_{i+1} - r_i},
    \label{eq:opacity_interp}
\end{equation}
where $r_i$ and $r_{i+1}$ are the resolutions of quality levels $L_i$ and $L_{i+1}$, respectively, and $r_i \leq r_t < r_{i+1}$.

Our interpolation scheme ensures smooth transitions between different levels of detail in our model.
We illustrate the renderings of the interpolated models between $L_0$ and $L_1$ in \cref{fig:smooth_transition}.

Additionally, by dynamically and continuously adjusting the influence of splats based on their opacity, {\name} could allow for efficient view-adaptive rendering, which optimizes rendering performance by adjusting the quality levels based on the viewer’s gaze and the distance to objects. Detailed analysis and comparison, along with illustrative figures, can be found in the supplementary materials.

%% file: sec/5_experiments.tex
\begin{table*}[t]
    \centering
    \caption{Quantitative comparison results on synthetic Blender dataset~\cite{mildenhall2021nerf} at different quality levels. The model size is normalized.}
    \label{tab:synthetic_result}
    \vspace{-0.3cm}
    \resizebox{\linewidth}{!} {%
        \begin{tabular}{c|ccc|ccc|ccc|ccc}
            \hline 
            \multirow{2}{*}{ Method } & \multicolumn{3}{c|}{$L_0$} & \multicolumn{3}{c|}{$L_1$} & \multicolumn{3}{c|}{$L_2$} & \multicolumn{3}{c}{$L_3$} \\
            %
             & SSIM $\uparrow$ & LPIPS $\downarrow$ & Size $\downarrow$ & SSIM $\uparrow$ & LPIPS $\downarrow$ & Size $\downarrow$ & SSIM $\uparrow$ & LPIPS $\downarrow$ & Size $\downarrow$ & SSIM $\uparrow$ & LPIPS $\downarrow$ & Size $\downarrow$ \\
            \hline 
            Downsample & 0.827 & 0.117 & 0.127 & 0.875 & 0.092 & 0.141 & 0.907 & 0.069 & 0.252 & 0.929 & 0.060 & 0.430 \\
            Single & 0.747 & 0.131 & 1.000 & 0.864 & 0.070 & 1.000 & 0.946 & 0.032 & 1.000 & 0.969 & 0.027 & 1.000 \\
            \hline
            Multiscale & \textbf{0.984} & \textbf{0.014} & \textbf{0.127} & 0.980 & 0.016 & 0.272 & \textbf{0.976} & \textbf{0.018} & 0.540 & 0.969 & \textbf{0.027} & 1.000 \\
            \name & \textbf{0.984} & \textbf{0.014} & \textbf{0.127} & \textbf{0.981} & \textbf{0.015} & \textbf{0.141} & 0.970 & 0.026 & \textbf{0.252} & \textbf{0.970} & 0.044 & \textbf{0.430} \\
            $\Delta$ & - & - & - & 0.001 & -0.001 & -92.91\% & -0.006 & 0.008 & -114.29\% & 0.001 & 0.017 & -132.56\% \\
            \hline
        \end{tabular}
    }
    \vspace{0.2cm}
    \centering
    \caption{Quantitative comparison results on Mip-NeRF360 dataset~\cite{barron2022mip} at different quality levels. The model size is normalized.}
    \label{tab:360_result}
    \vspace{-0.3cm}
    \resizebox{\linewidth}{!} {%
        \begin{tabular}{c|ccc|ccc|ccc|ccc}
            \hline 
            \multirow{2}{*}{ Method } & \multicolumn{3}{c|}{$L_0$} & \multicolumn{3}{c|}{$L_1$} & \multicolumn{3}{c|}{$L_2$} & \multicolumn{3}{c}{$L_3$} \\
            %
             & SSIM $\uparrow$ & LPIPS $\downarrow$ & Size $\downarrow$ & SSIM $\uparrow$ & LPIPS $\downarrow$ & Size $\downarrow$ & SSIM $\uparrow$ & LPIPS $\downarrow$ & Size $\downarrow$ & SSIM $\uparrow$ & LPIPS $\downarrow$ & Size $\downarrow$ \\
            \hline 
            Downsample & 0.548 & 0.314 & 0.239 & 0.678 & 0.236 & 0.413 & 0.778 & 0.194 & 0.520 & 0.870 & 0.166 & 0.568 \\
            Single & 0.635 & 0.201 & 1.000 & 0.752 & 0.164 & 1.000 & 0.881 & 0.119 & 1.000 & 0.918 & 0.124 & 1.000 \\
            \hline
            Multiscale & \textbf{0.957} & \textbf{0.052} & \textbf{0.239} & \textbf{0.947} & \textbf{0.065} & 0.531 & \textbf{0.928} & \textbf{0.096} & 0.783 & 0.918 & \textbf{0.124} & 1.000 \\
            \name & \textbf{0.957} & \textbf{0.052} & \textbf{0.239} & 0.936 & 0.080 & \textbf{0.413} & \textbf{0.928} & 0.111 & \textbf{0.520} & \textbf{0.925} & 0.161 & \textbf{0.568} \\
            $\Delta$ & - & - & - & -0.011 & 0.015 & -28.57\% & 0.000 & 0.015 & -50.58\% & 0.007 & 0.037 & -76.06\% \\
            \hline
        \end{tabular}
    }
    \vspace{0.2cm}
    \centering
    \caption{Quantitative comparison results on Tank\&Temples dataset~\cite{knapitsch2017tanks} at different quality levels. The model size is normalized.}
    \label{tab:tandt_result}
    \vspace{-0.3cm}
    \resizebox{\linewidth}{!} {%
        \begin{tabular}{c|ccc|ccc|ccc|ccc}
            \hline 
            \multirow{2}{*}{ Method } & \multicolumn{3}{c|}{$L_0$} & \multicolumn{3}{c|}{$L_1$} & \multicolumn{3}{c|}{$L_2$} & \multicolumn{3}{c}{$L_3$} \\
            %
             & SSIM $\uparrow$ & LPIPS $\downarrow$ & Size $\downarrow$ & SSIM $\uparrow$ & LPIPS $\downarrow$ & Size $\downarrow$ & SSIM $\uparrow$ & LPIPS $\downarrow$ & Size $\downarrow$ & SSIM $\uparrow$ & LPIPS $\downarrow$ & Size $\downarrow$ \\
            \hline 
            Downsample & 0.530 & 0.340 & 0.104 & 0.602 & 0.302 & 0.241 & 0.724 & 0.238 & 0.424 & 0.868 & 0.154 & 0.608 \\
            Single & 0.640 & 0.198 & 1.000 & 0.764 & 0.140 & 1.000 & 0.885 & 0.092 & 1.000 & 0.923 & 0.106 & 1.000 \\
            \hline
            Multiscale & \textbf{0.958} & \textbf{0.051} & \textbf{0.104} & \textbf{0.946} & \textbf{0.060} & 0.272 & 0.934 & \textbf{0.077} & 0.528 & \textbf{0.923} & \textbf{0.106} & 1.000 \\
            \name & \textbf{0.958} & \textbf{0.051} & \textbf{0.104} & 0.942 & 0.076 & \textbf{0.241} & \textbf{0.935} & 0.103 & \textbf{0.424} & 0.916 & 0.163 & \textbf{0.608} \\
            $\Delta$ & - & - & - & -0.004 & 0.016 & -12.86\% & 0.001 & 0.026 & -24.53\% & -0.007 & 0.057 & -64.47\% \\
            \hline
        \end{tabular}
    }
    \vspace{0.2cm}
    \centering
    \caption{Quantitative comparison results on Deep Blending dataset~\cite{hedman2018deep} at different quality levels. The model size is normalized.}
    \label{tab:db_result}
    \vspace{-0.3cm}
    \resizebox{\linewidth}{!} {%
        \begin{tabular}{c|ccc|ccc|ccc|ccc}
            \hline 
            \multirow{2}{*}{ Method } & \multicolumn{3}{c|}{$L_0$} & \multicolumn{3}{c|}{$L_1$} & \multicolumn{3}{c|}{$L_2$} & \multicolumn{3}{c}{$L_3$} \\
            %
             & SSIM $\uparrow$ & LPIPS $\downarrow$ & Size $\downarrow$ & SSIM $\uparrow$ & LPIPS $\downarrow$ & Size $\downarrow$ & SSIM $\uparrow$ & LPIPS $\downarrow$ & Size $\downarrow$ & SSIM $\uparrow$ & LPIPS $\downarrow$ & Size $\downarrow$ \\
            \hline 
            Downsample & 0.820 & 0.168 & 0.319 & 0.880 & 0.132 & 0.431 & 0.909 & 0.157 & 0.489 & 0.913 & 0.236 & 0.544 \\
            Single & 0.829 & 0.152 & 1.000 & 0.910 & 0.095 & 1.000 & 0.959 & 0.094 & 1.000 & 0.959 & 0.171 & 1.000 \\
            \hline
            Multiscale & \textbf{0.966} & \textbf{0.045} & \textbf{0.319} & \textbf{0.967} & \textbf{0.055} & 0.611 & \textbf{0.967} & \textbf{0.077} & 0.825 & \textbf{0.959} & \textbf{0.171} & 1.000 \\
            \name & \textbf{0.966} & \textbf{0.045} & \textbf{0.319} & 0.961 & 0.097 & \textbf{0.431} & 0.954 & 0.125 & \textbf{0.489} & 0.951 & 0.219 & \textbf{0.544} \\
            $\Delta$ & - & - & - & -0.006 & 0.042 & -41.76\% & -0.013 & 0.048 & -68.71\% & -0.008 & 0.048 & -83.82\% \\
            \hline
        \end{tabular}
    }
\end{table*}

\section{Experiments} \label{sec:exp}


In this section, we first detail the experimental setup of {\name} in Section \ref{subsec:setup}.
We then evaluate the performance of our approach in Section \ref{subsec:results}.
We also conduct ablation studies to analyze the contributions of key components within our method.

\subsection{Experimental Setup} \label{subsec:setup}

\textbf{Dataset}.
We evaluated our method using $19$ objects and real scenes from various datasets, including Synthetic Blender \cite{mildenhall2021nerf}, Mip-NeRF360 \cite{barron2022mip}, Tanks\&Temples \cite{knapitsch2017tanks}, and Deep Blending \cite{hedman2018deep}.
These datasets encompass a diverse range of object-centric, indoor, and outdoor scenes, providing a robust basis for testing.
In addition to full-scale images, we down-scaled each dataset by factors of 2$\times$, 4$\times$, and 8$\times$, reducing the image resolution by half, a quarter, and one-eighth in each direction, respectively. 
This provided multiple scales for constructing the layered representation.

\textbf{Implementation}.
Our implementation is based on the official release of the 3D Gaussian Splatting code~\cite{gaussiancode}.
We initially trained the base layer from the dataset with the lowest resolution.
In subsequent training stages, we fixed the parameters of Gaussian splats in prior layers, except for their opacity, and trained the enhancement layer and the opacity of prior layers on datasets with the corresponding quality levels. 
All hyperparameters remained consistent across training stages.
The training process was conducted on an NVIDIA A100 GPU.

\textbf{Comparison Method}.
We compare {\name} with three alternative approaches: 
\begin{itemize}
    \item \textit{Single}. A full-resolution model is trained and then rendered at multiple scales.
    This approach demonstrates the trade-offs in scalability and quality when a single model is tasked with adapting to different resolutions without specific optimization for each scale.
    \item \textit{Multiscale}. Separate models are trained independently for each scale (1$\times$, 2$\times$, 4$\times$, and 8$\times$), resulting in four distinct models that form a multiscale representation.
    This is similar to tree-based hierarchies where each level is independently represented shown in \cref{fig:hierarchy_illus}, and serves as the upper bound for reconstruction quality.
    \item \textit{Downsample}. Following Fan \etal~\cite{fan2023lightgaussian}, this method downsamples the full-resolution model by calculating significance scores based on the opacity and scale of each splat, removing those with lower scores.
    This approach serves as the lower bound and highlights the limitations of using traditional point cloud down- and up-sampling techniques~\cite{shi2023enabling,zhang2022yuzu} to construct an LOD representation for 3DGS which is a point-based representation. 
\end{itemize}

\textbf{Metrics}. We evaluate the visual quality using SSIM and LPIPS.
We exclude PSNR from our evaluation as it primarily estimates pixel-wise error, which makes it less sensitive to blurriness and low-resolution artifacts~\cite{wang2004image,zhang2024fregs}, as discussed in \cref{subsec:progressive_3dgs}.
In addition to quality, we compare the model sizes across different methods. 
Given that 3DGS models vary significantly in size depending on the complexity of the 3D content they represent, we normalize the model sizes to a range of $\left(0, 1\right]$ by dividing each model's size by the maximum model size within that 3D content.

\begin{figure*}[t]%
    \centering
    \begin{subfigure}{0.245\linewidth}
        \includegraphics[width=\linewidth]{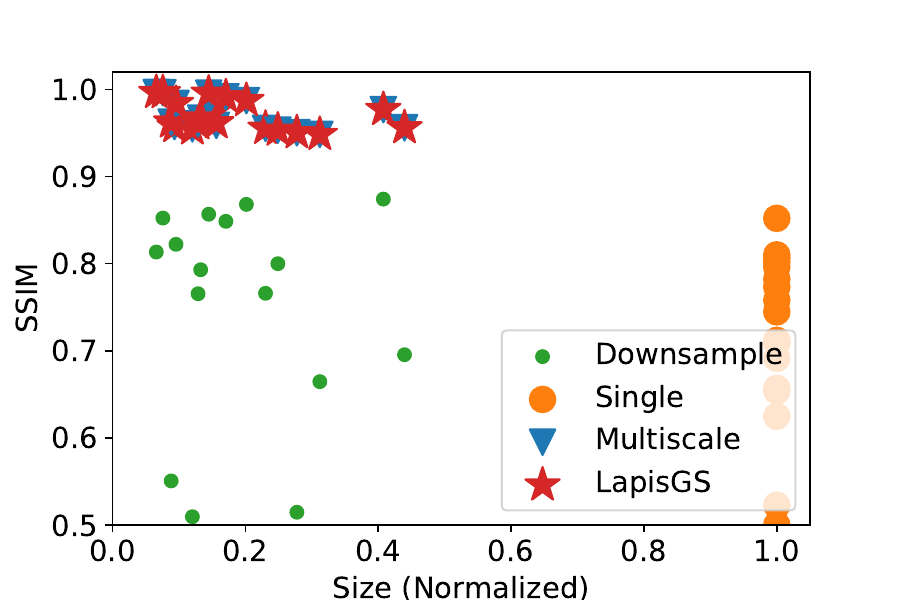}
        \caption{$L_0$.}
    \end{subfigure}
    \begin{subfigure}{0.245\linewidth}
        \includegraphics[width=\linewidth]{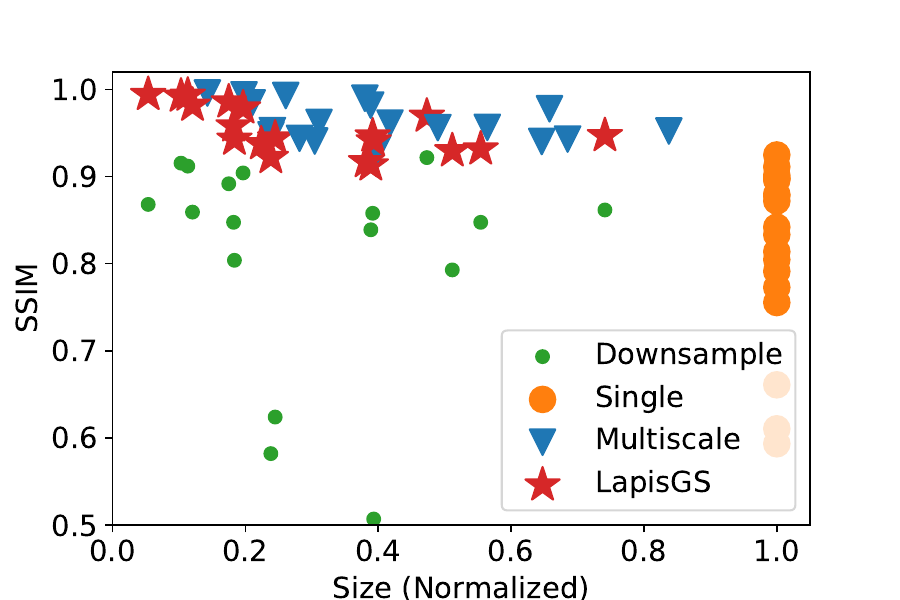}
        \caption{$L_1$.}
    \end{subfigure}
    \begin{subfigure}{0.245\linewidth}
        \includegraphics[width=\linewidth]{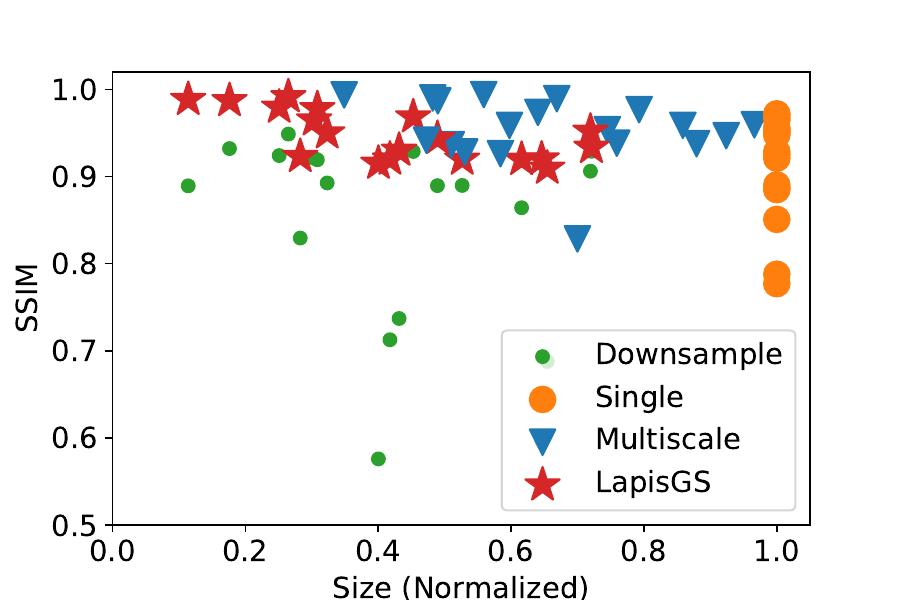}
        \caption{$L_2$.}
    \end{subfigure}
    \begin{subfigure}{0.245\linewidth}
        \includegraphics[width=\linewidth]{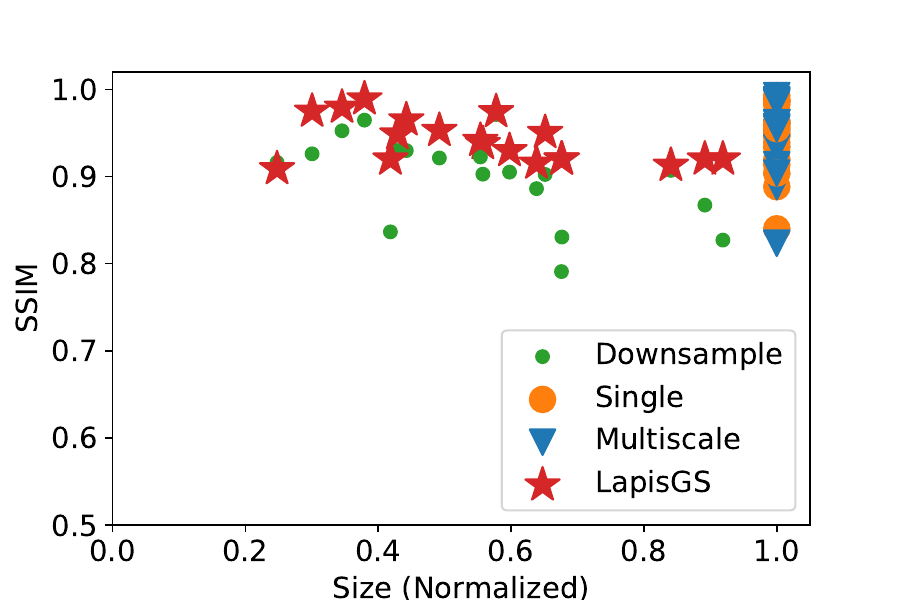}
        \caption{$L_3$.}
    \end{subfigure}
    \caption{Each point represents the overall quality of a scene/object with the corresponding model size at a given quality level. As the level increases, {\name} achieves high visual quality \sg{(SSIM)} with a considerably smaller model size. 
    }%
    \label{fig:scatter}
\end{figure*}


\subsection{Results and Evaluation} \label{subsec:results}

\textbf{Quantitative Comparison}. 
We present the results from four datasets in \crefrange{tab:synthetic_result}{tab:db_result}. As our comparison focuses on the trade-off between model size and rendering quality, the size and quality difference and change in size of our method compared to the Multiscale method is also presented. 
Note that we quantify the size change relative to the new model size, which expresses the reduction as a negative percentage. A more negative value indicates a greater size decrease.
To better illustrate the comparison, we also show the visual quality against normalized model size for each scene and object at each resolution scale, in \cref{fig:scatter}.
Several key observations are highlighted.

First, the downsampled model shows significant degradation in visual quality as model size decreases.
This suggests that traditional downsampling methods for point clouds are inadequate for 3DGS, primarily because they fail to re-optimize or re-learn the model for the reduced set of splats. Given the unique distribution and anisotropic nature of Gaussian splats, simply reducing the number of splats without retraining does not capture the high-frequency details or maintain the model’s original fidelity.

Second, {\name} consistently achieves the smallest model size across all datasets and scales, demonstrating the effectiveness of our progressive multiscale training pipeline:
by enabling feature sharing across different LODs and multi-level optimization and occupancy maps, our approach significantly compacts the representation.
This reduction in model size directly correlates to decreased rendering times, further illustrating the efficiency of our layered progressive model in adaptive rendering scenarios.

Third, {\name} achieves notable improvements in both visual quality and model size compared to the single-scale model.
At lower resolutions, our approach yields enhancements of up to 50.71\% in SSIM, 286.53\% in LPIPS with 23\% of the original model size.
These results emphasize the necessity of constructing a multiscale 3DGS LOD model and highlight the effectiveness and efficiency of our method in maintaining visual fidelity while reducing data overhead.

Finally, {\name} rivals the multiscale model, which serves as the upper bound for reconstruction quality, achieving comparable visual fidelity with up to a 2.33$\times$ smaller model size.
This demonstrates that our method effectively balances high and low resolutions, creating a compact yet highly detailed representation.

In summary, our layered progressive model ensures high visual quality with a compact model that supports efficient and adaptive rendering, making it an effective solution for adaptive 3DGS streaming.

\begin{figure*}[t]
    \centering
    \includegraphics[width=1.0\linewidth]{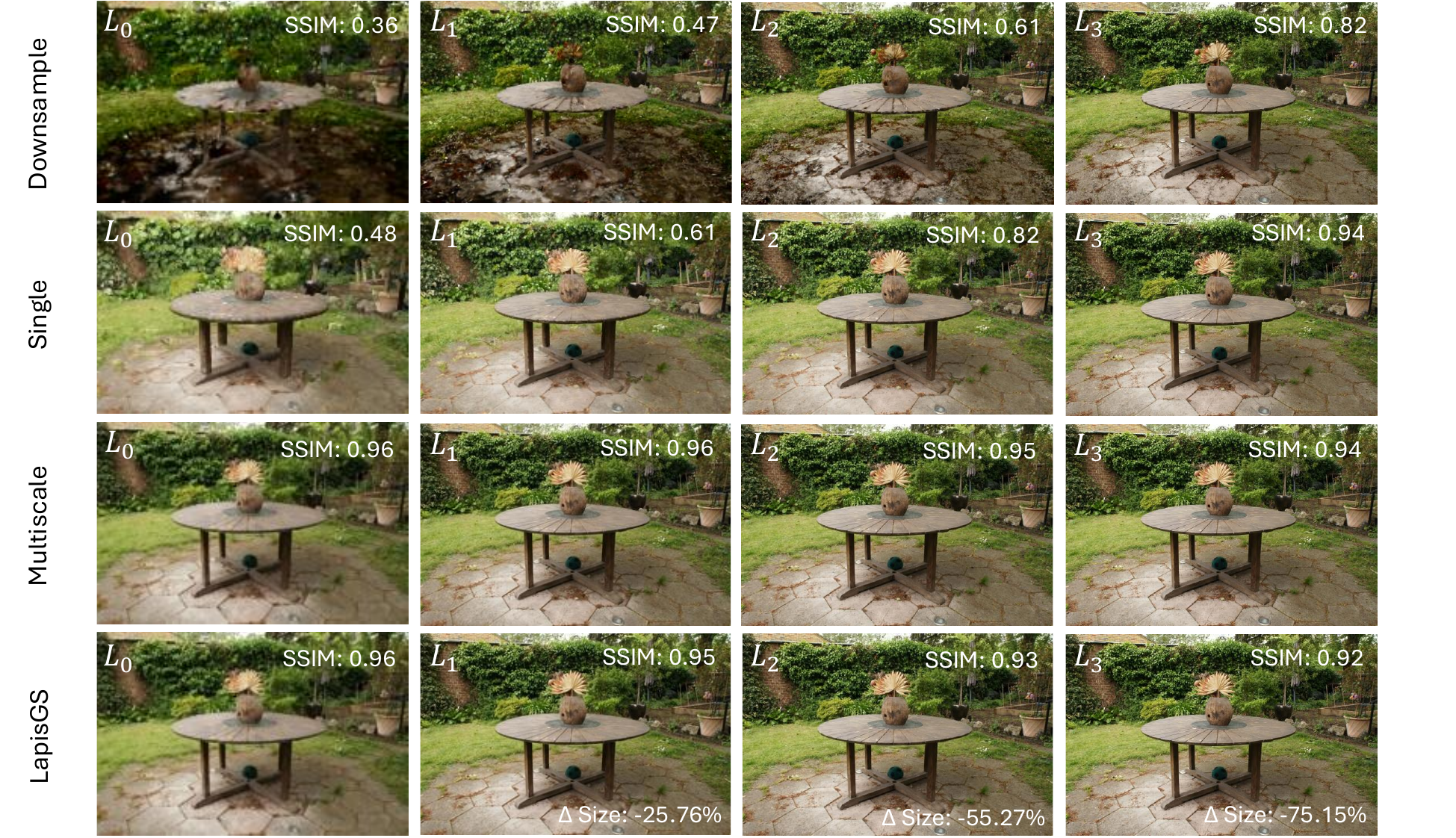}
    \caption{Sample renderings of \textit{Garden} at different levels.
    {\name} eliminates the usual artifacts seen in downsampled and single-scale models, achieving comparable quality to the Multiscale method with considerably lower data size, as shown in \cref{tab:360_result}.}
    \label{fig:garden}
\end{figure*}

\textbf{Qualitative Comparison}.
We present the qualitative comparison with the other methods in \cref{fig:garden}. 
More detailed qualitative comparisons are shown in the supplementary materials due to the page limit.
As observed, {\name} successfully avoids the common artifacts seen in downsampled and single-scale models, achieving results comparable to the Multiscale method but with significantly reduced model size.
This improvement highlights the effectiveness of our layered progressive model in capturing fine-grained features while maintaining a compact model size, making it an ideal choice for adaptive 3DGS streaming.

\begin{table}[htbp]
    \centering
    \caption{The average size and quality difference in percentage of {\name} at different scales, compared to the Freeze method.}
        \begin{tabular}{c|ccc}
            \hline
             Dataset & $\Delta$ SSIM $\uparrow$ & $\Delta$ LPIPS $\downarrow$ & $\Delta$ Size $\downarrow$\\
            \hline 
            Synthetic~\cite{mildenhall2021nerf} & 0.82\% & -26.74\% & -118.90\% \\
            360~\cite{barron2022mip} & 4.61\% & -22.34\% & -142.19\% \\
            T\&T~\cite{knapitsch2017tanks} & 5.01\% & -32.31\% & -123.10\% \\
            DB~\cite{hedman2018deep} & 1.71\% & -12.28\% & -201.46\% \\
            \hline
        \end{tabular}
    \label{tab:ablation}
\end{table}

\textbf{Ablation Study}.
To build a layered feature-sharing LOD model, we chose to optimize only the opacity values of the previous layers and freeze other parameters to balance efficient layer integration and visual coherence.
This step empowers the representation compaction and adjustment, as well as view-adaptive rendering, by dynamically determining which splats contribute most significantly to the visual fidelity at various distances from the viewpoint.

To explore the effect of the proposed multi-level optimization approach, we conducted an ablation study.
Specifically, we evaluated a variant of our method where all parameters, including opacity, are frozen when training enhancement layers.
This method, denoted as ``Freeze'', serves as a baseline to highlight the importance of dynamic opacity optimization.
We evaluated the visual quality and model size on all four datasets, comparing our method against the Freeze method.

We present the average results in \cref{tab:ablation} and per-scene quantitative and qualitative results in the supplementary. The results highlight that without opacity optimization, model size increases significantly, and visual quality deteriorates.
The visual quality degrades because lower-layer splats, which cannot capture high-frequency features, continue to influence higher-layer representations.
Consequently, the model size increases significantly as additional splats are added to compensate for the deficiencies of these lower-layer splats, leading to redundancy.
In contrast, our method achieves more compact models and superior visual fidelity by efficiently pruning less significant splats. 

Overall, {\name} ensures a balance between model size and rendering quality, demonstrating the effectiveness of dynamic opacity optimization for scalable 3DGS.

%% file: sec/6_conclusion.tex
\section{Discussion and Conclusion}
\label{sec:conclusion}

In this paper, we introduced {\name}, a layered progressive 3DGS designed for adaptive streaming.
Our approach addresses key challenges in 3D content streaming by optimizing the balance between visual quality and model size.
We leverage visual information sharing across multiple LODs, which is crucial for real-time rendering and adaptive streaming where resources are limited.
A key advantage of {\name} is its progressive nature, where each layer builds upon the previous one, allowing for efficient and coherent integration of details. 
By selectively optimizing the opacity of splats in lower layers, we reduce redundancy and ensure that only relevant splats enhance visual fidelity, particularly for high-frequency details.
This progressive framework avoids the inefficiencies of traditional models that either downsample or train separate models for each scale.
Our method demonstrates superior efficiency and flexibility, achieving substantial improvements in visual quality and model compactness, as evidenced by high SSIM and low LPIPS scores across various datasets.
Additionally, it supports view-adaptive streaming and rendering through dynamic pruning and interpolation of opacity based on network conditions, ensuring smooth transitions between quality levels and a seamless user experience. 
For a complementary analysis and comparison, including visual illustrations, we direct readers to the supplementary materials.

While our method performs well, there are areas for improvement.
First, our approach is optimized for static scenes. Extending the model to handle dynamic elements or real-time updates in changing environments is an important direction for future research, especially for more interactive applications.
Second, a notable aspect not explored in this work is the evaluation of performance under fluctuating network conditions.
These tasks are beyond the scope of this paper, which focuses on developing an adaptive streaming model. 
However, we refer readers to our recent work~\cite{sun2025tsla}, where we develop, implement, and evaluate the first DASH-based dynamic 3DGS streaming system built on our {\name}. This system demonstrates superior performance in both live and on-demand streaming.

%% file: sec/X_suppl.tex
\clearpage
\setcounter{page}{1}
\maketitlesupplementary

\section{View-Adaptive Rendering} \label{sec:adaptive}

\begin{figure}[t]%
    \centering
    \begin{subfigure}{0.49\linewidth}
        \includegraphics[width=\linewidth]{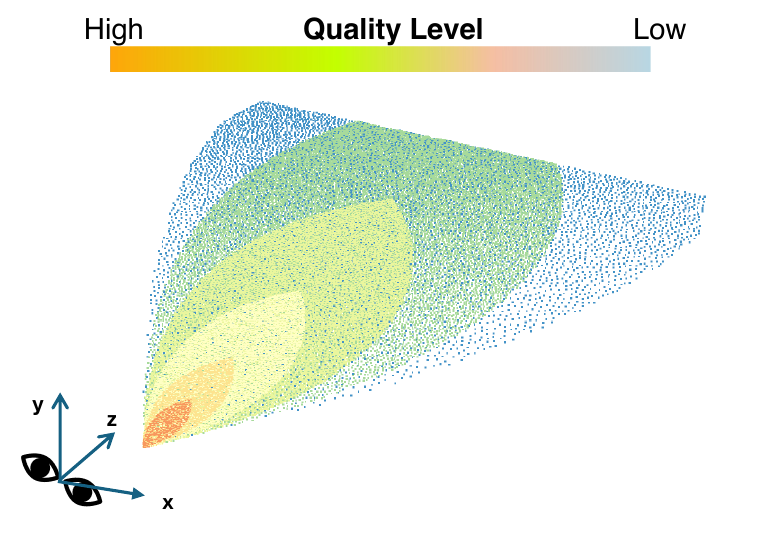}
        \caption{Discrete quality adaptation.}
        \label{subfig:discrete_rendering_illustrate}
    \end{subfigure}
    \begin{subfigure}{0.49\linewidth}
        \includegraphics[width=\linewidth]{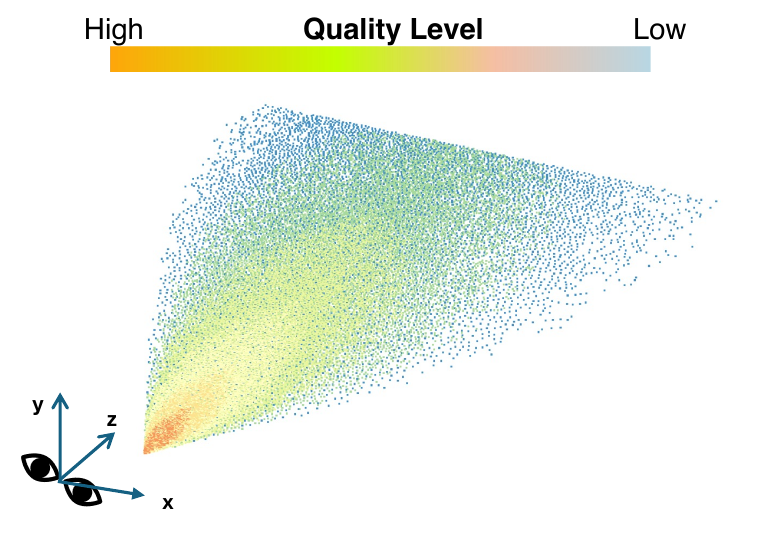}
        \caption{Continuous quality adaptation.}
        \label{subfig:continuous_rendering_illustrate}
    \end{subfigure}
    \caption{Illustration of discrete and continuous quality adaptation.}%
    \label{fig:rendering_illustrate}
\end{figure}

Distance-aware foveated rendering is a technique that optimizes rendering performance by adjusting the LOD in a 3D scene based on the viewer’s gaze and the distance to objects.
This method focuses computational resources on rendering high-quality images in the viewer's direct line of sight (foveal region), while reducing detail in peripheral areas, as shown in \cref{fig:rendering_illustrate}.

Space-based hierarchical structures commonly used in 3D rendering present challenges due to their discrete nature~\cite{kerbl2024hierarchical,ren2024octree,jiang20243dgs}.
The structure often results in chunk-based representation, which can lead to inefficient alignment with arbitrarily oriented scene elements such as walls, pillars, or stairs.
This misalignment causes the detailed rendering of the scene center to be less effective~\cite{schutz2019real}.
Although smooth transitions can be achieved within individual chunks, the chunk-wise rendering approach leads to popping artifacts, resulting in a fundamentally discrete quality transition, as shown in \cref{subfig:discrete_rendering_illustrate}.

\begin{figure}[t]%
    \centering
    \begin{subfigure}{0.49\linewidth}
        \includegraphics[width=\linewidth]{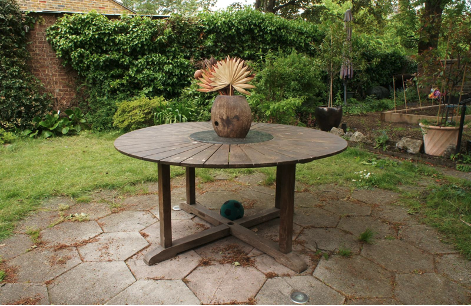}
        \caption{Ground truth rendering.}
    \end{subfigure}
    \begin{subfigure}{0.49\linewidth}
        \includegraphics[width=\linewidth]{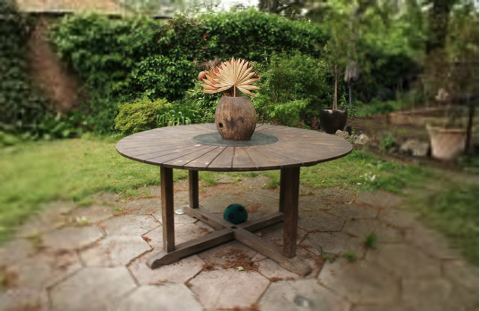}
        \caption{Continuous adaptive rendering.}
    \end{subfigure}
    \caption{The sample rendering results of \textit{Garden}.}%
    \label{fig:adaptive_rendeirng_example}
\end{figure}

Our proposed layered structure addresses these limitations by using a cumulative stacking approach, where layers progressively merge to form higher-quality levels.
This architecture allows for the sharing of visual information across different layers, making it naturally suitable for adaptive quality rendering.
Unlike the chunk-wise approach of tree structures, our method evaluates and adjusts the opacity of Gaussian splats on a splat-wise basis with \cref{eq:opacity_interp}, enabling smoother, continuous transitions.
As illustrated in \cref{subfig:continuous_rendering_illustrate}, our method supports view-adaptive rendering, providing seamless quality transitions and enhanced visual fidelity.
We also show an example of view-adaptive rendering from our model in \cref{fig:adaptive_rendeirng_example}.

\section{Qualitative Results}

\begin{figure*}[t]
    \centering
    \includegraphics[width=\linewidth]{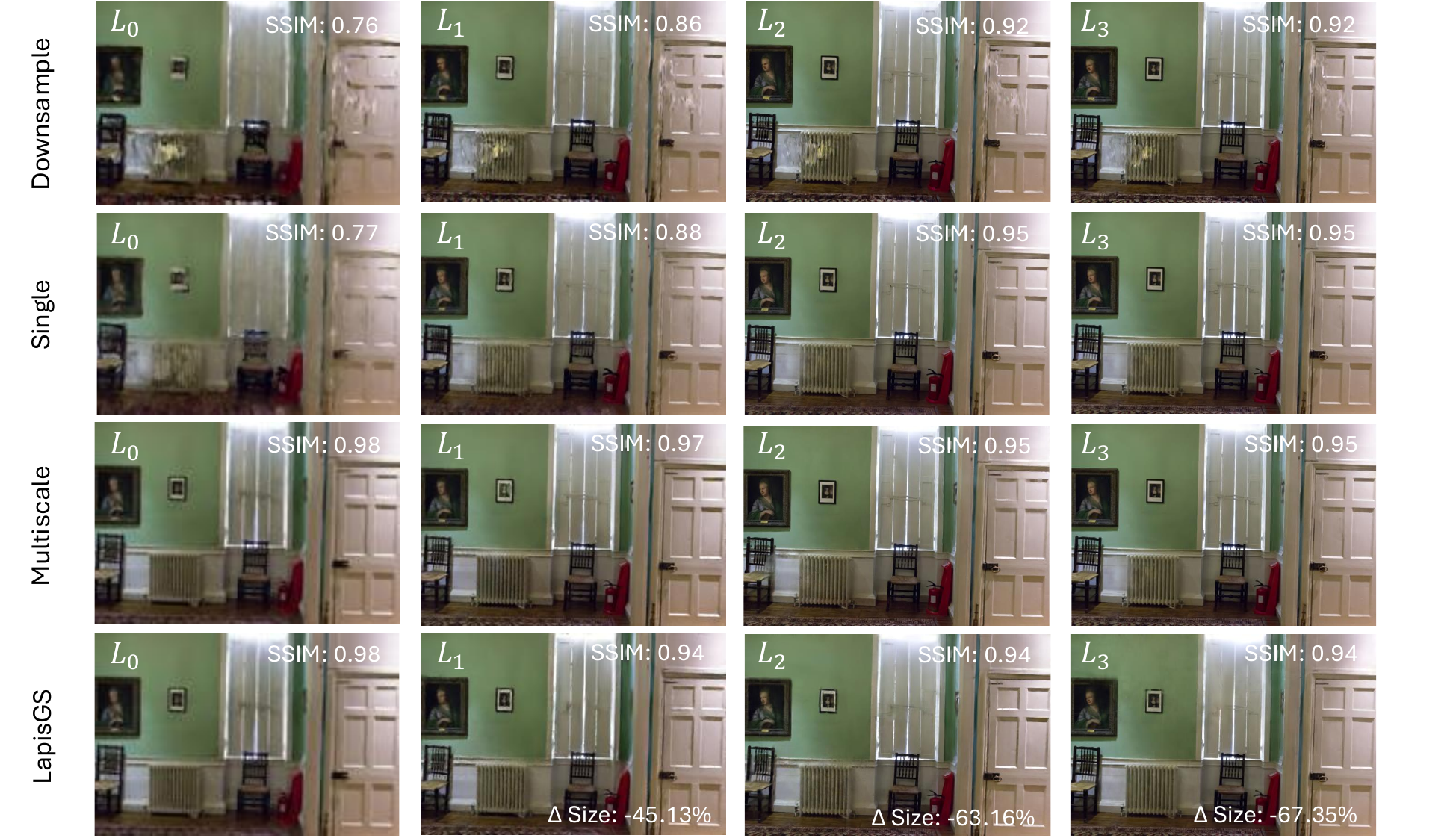}
    \caption{Sample renderings of \textit{Drjohnson} from Deep Blending dataset\cite{hedman2018deep} at different scales.}
    \label{fig:drjohnson}
\end{figure*}

\begin{figure*}[t]
    \centering
    \includegraphics[width=\linewidth]{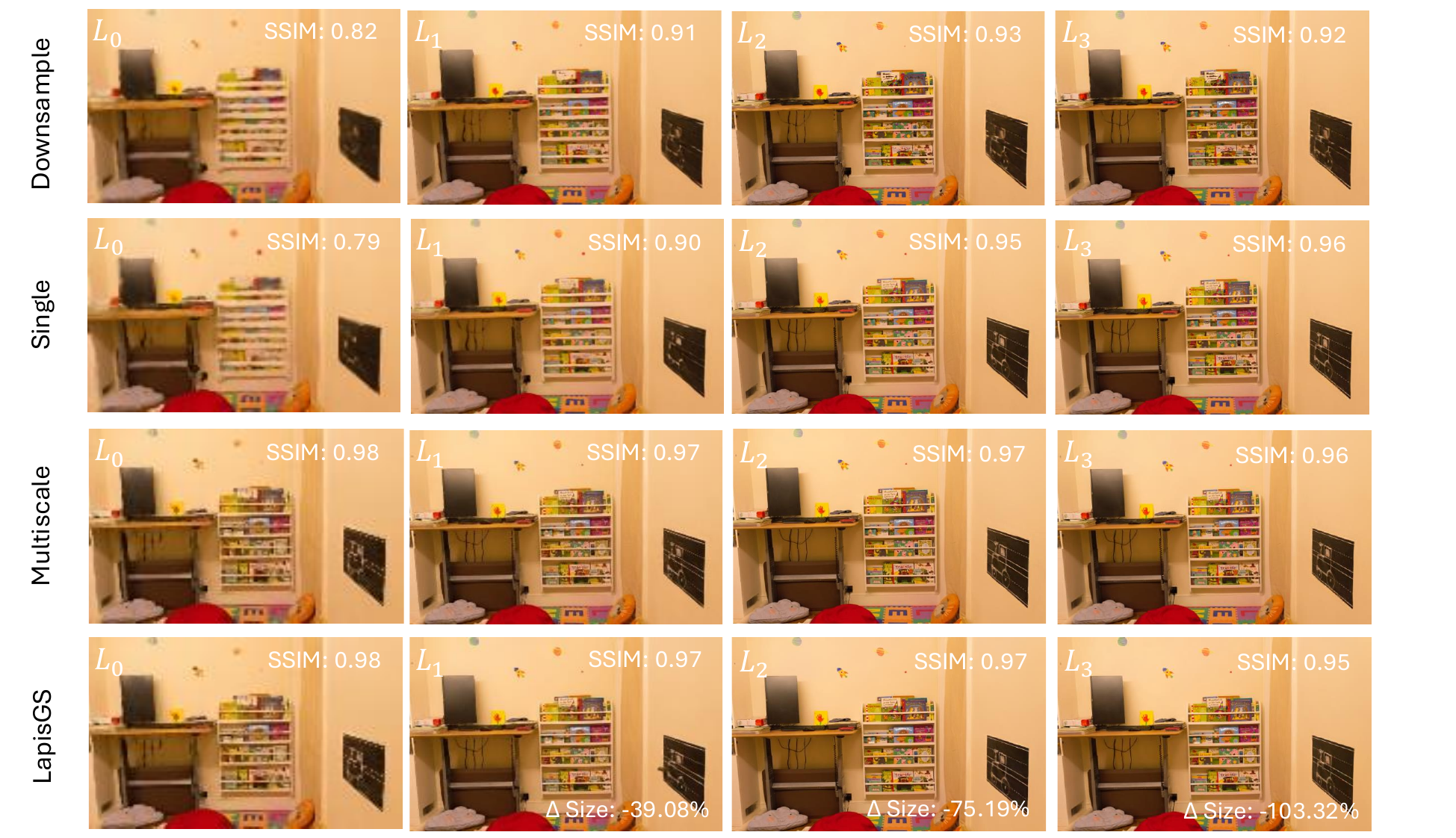}
    \caption{Sample renderings of \textit{Playroom} from Deep Blending dataset\cite{hedman2018deep} at different scales.}
    \label{fig:playroom}
\end{figure*}

\begin{figure*}[t]
    \centering
    \includegraphics[width=\linewidth]{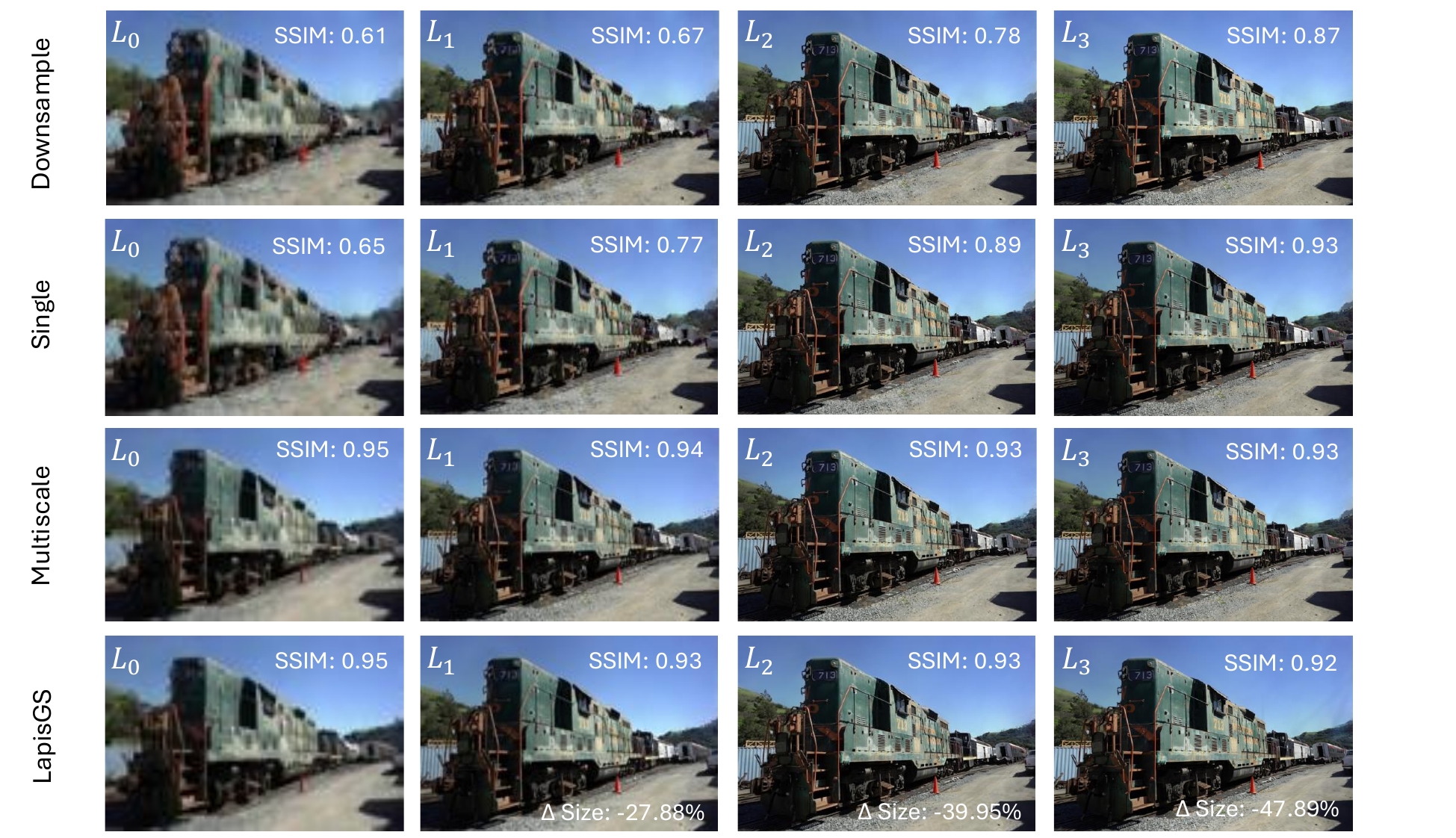}
    \caption{Sample renderings of \textit{Train} from Tank\&Temples dataset\cite{knapitsch2017tanks} at different scales.}
    \label{fig:train}
\end{figure*}

\begin{figure*}[t]
    \centering
    \includegraphics[width=\linewidth]{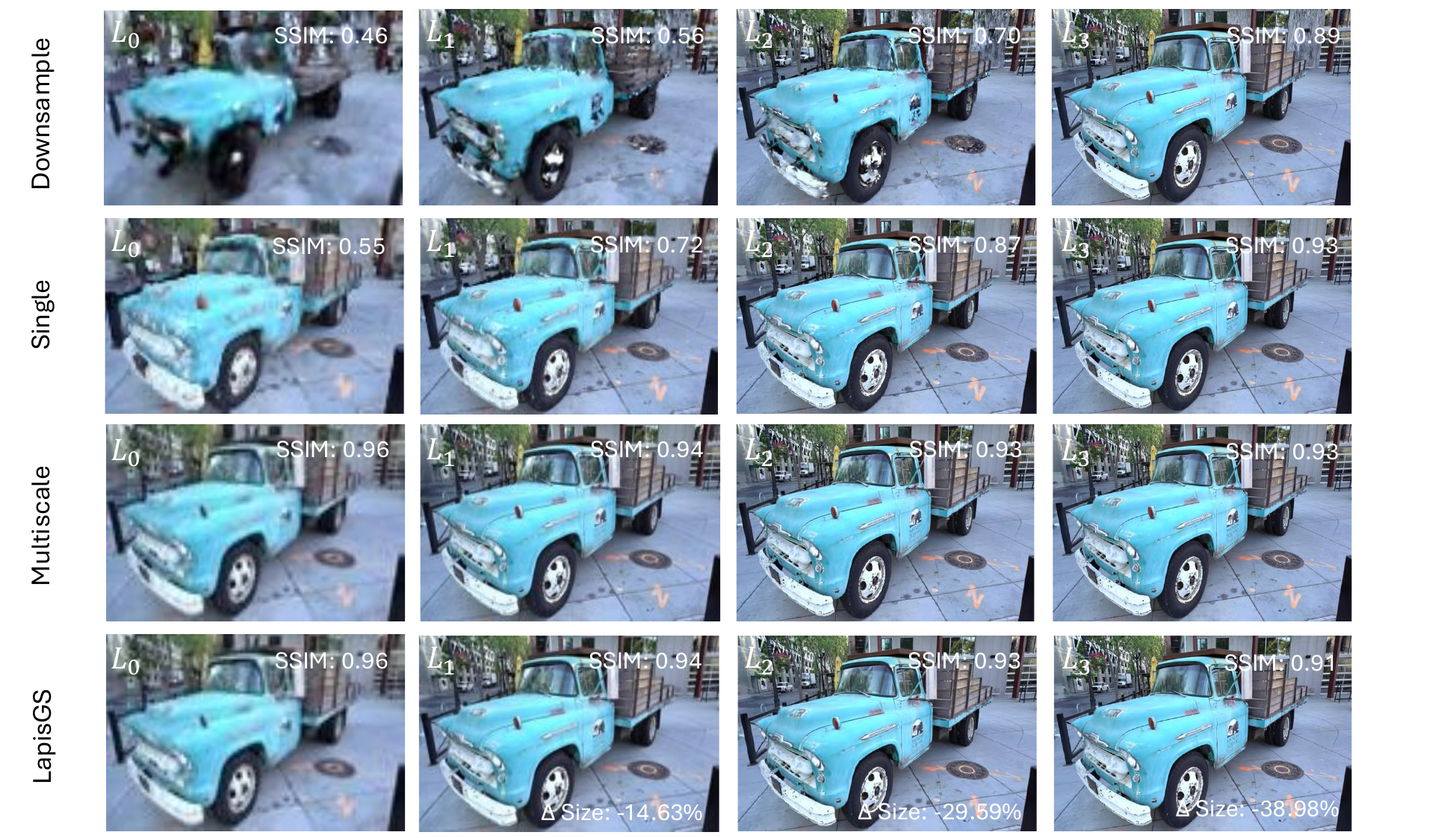}
    \caption{Sample renderings of \textit{Truck} from Tank\&Temples dataset\cite{knapitsch2017tanks} at different scales.}
    \label{fig:truck}
\end{figure*}

\begin{figure*}[t]
    \centering
    \includegraphics[width=\linewidth]{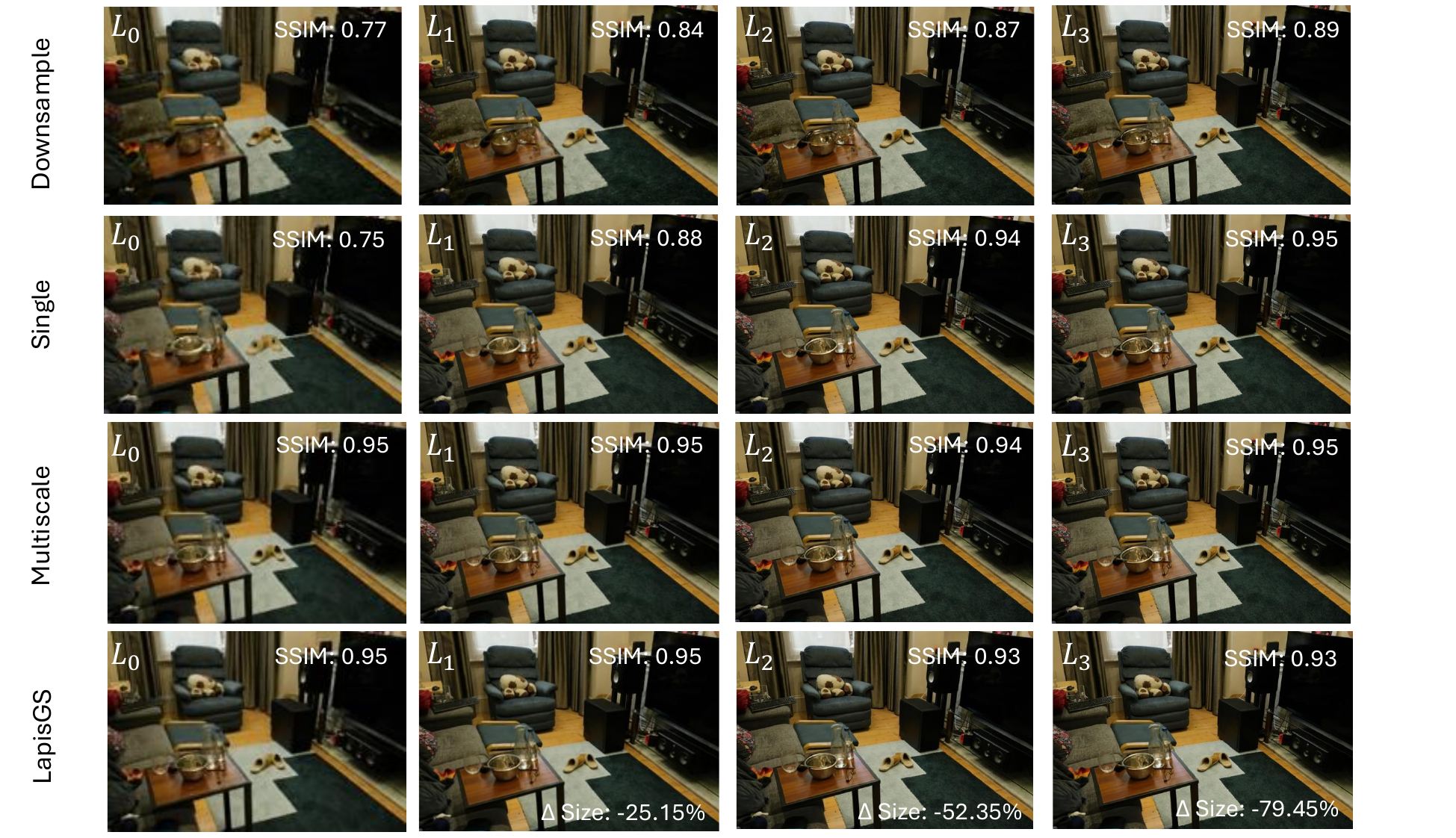}
    \caption{Sample renderings of \textit{Room} from Mip-NeRF360 dataset~\cite{barron2022mip} at different scales.}
    \label{fig:room}
\end{figure*}

\begin{figure*}[t]
    \centering
    \includegraphics[width=\linewidth]{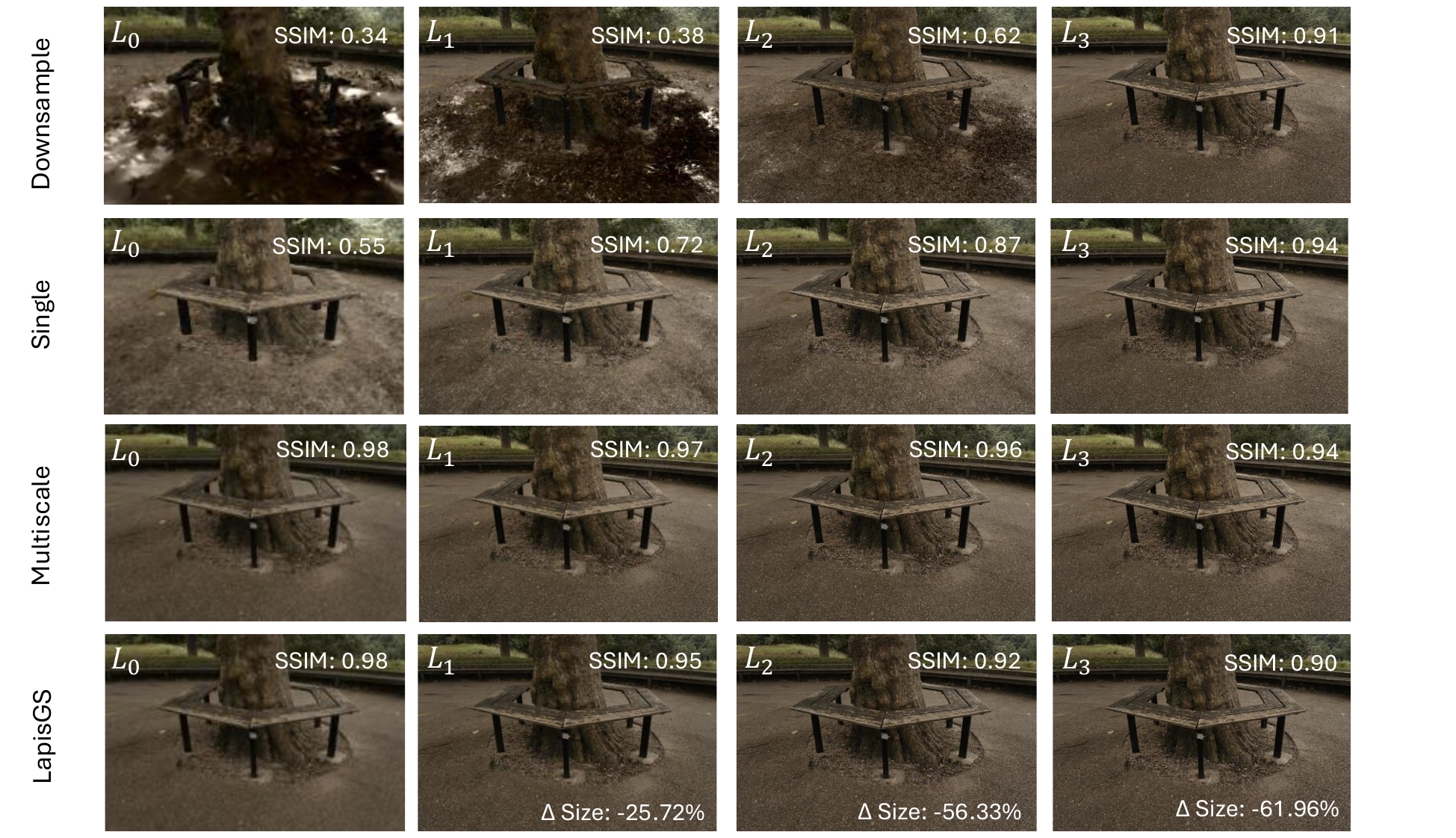}
    \caption{Sample renderings of \textit{Treehill} from Mip-NeRF360 dataset~\cite{barron2022mip} at different scales.}
    \label{fig:treehill}
\end{figure*}

We show the qualitative results of {\name} alongside comparison methods on a variety of scenes, including \textit{Drjohnson} and \textit{Playroom} in Deep Blending dataset~\cite{hedman2018deep}, \textit{Train} and \textit{Truck} in Tank\&Temples dataset~\cite{knapitsch2017tanks}, and \textit{Room} and \textit{Treehill} in Mip-NeRF360 dataset~\cite{barron2022mip}, as shown in \crefrange{fig:drjohnson}{fig:treehill}.

As observed, {\name} demonstrates superior performance in preserving intricate scene details while eliminating common rendering artifacts across various environments. Our method matches the visual quality of the Multiscale approach, which is the upper bound of reconstruction quality, while achieving a substantially reduced computational footprint.

\section{Per-Scene Quantitative and Qualitative Ablation Study}

\begin{figure*}[t]
    \centering
    \includegraphics[width=\linewidth]{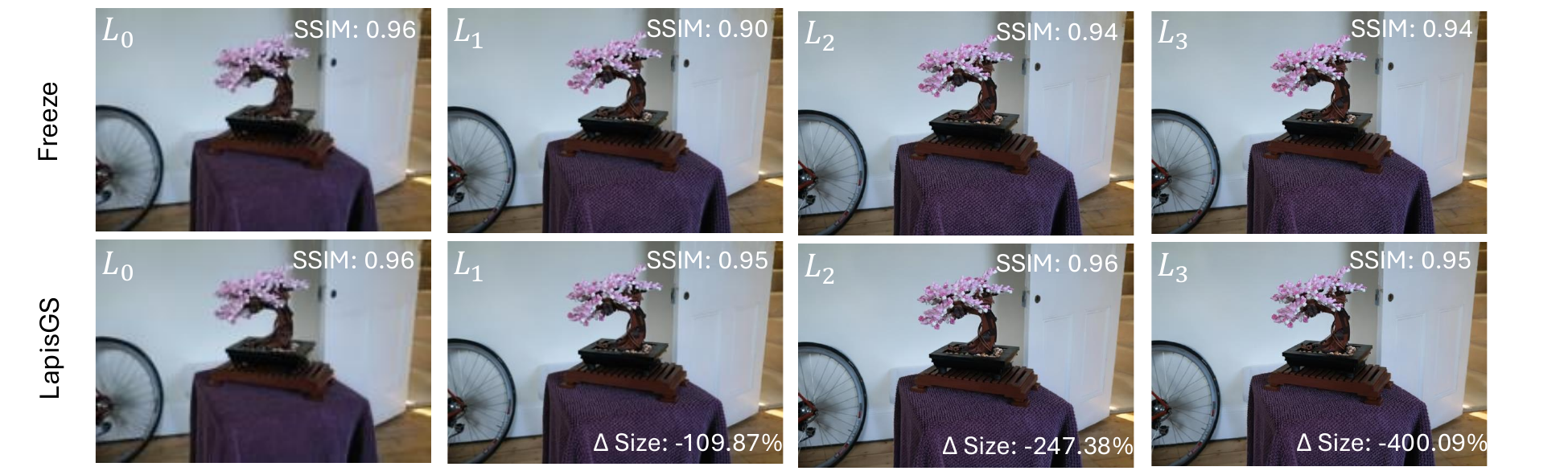}
    \caption{Sample renderings of \textit{Bonsai} from Mip-NeRF360 dataset\cite{barron2022mip} at different scales.}
    \label{fig:ablation_bonsai}
\end{figure*}

\begin{figure*}[t]
    \centering
    \includegraphics[width=\linewidth]{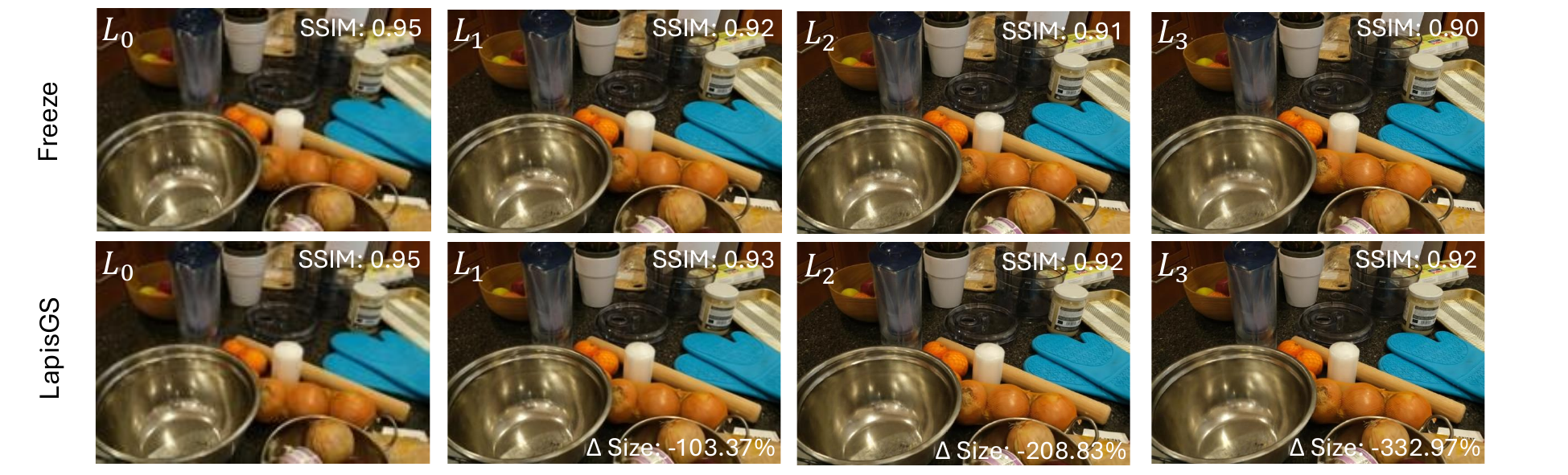}
    \caption{Sample renderings of \textit{Counter} from Mip-NeRF360 dataset\cite{barron2022mip} at different scales.}
    \label{fig:ablation_counter}
\end{figure*}

\begin{figure*}[t]
    \centering
    \includegraphics[width=\linewidth]{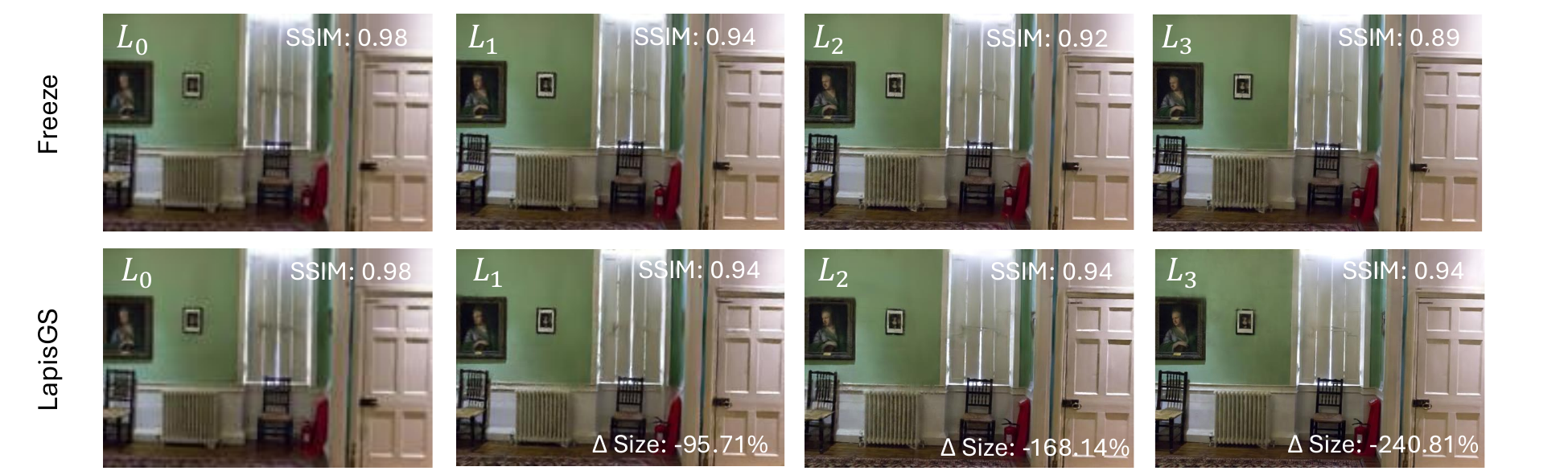}
    \caption{Sample renderings of \textit{Drjohnson} from Deep Blending dataset\cite{hedman2018deep} at different scales.}
    \label{fig:ablation_drjohnson}
\end{figure*}

\begin{figure*}[t]
    \centering
    \includegraphics[width=\linewidth]{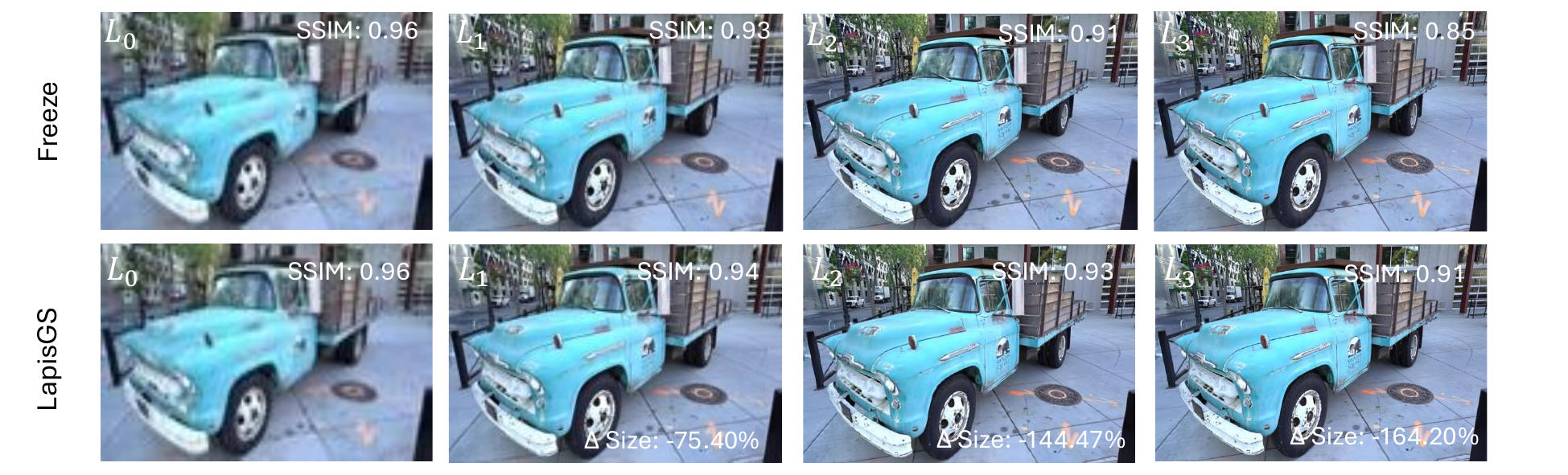}
    \caption{Sample renderings of \textit{Truck} from Tank\&Temples dataset\cite{knapitsch2017tanks} at different scales.}
    \label{fig:ablation_truck}
\end{figure*}

To provide a comprehensive comparison of the effectiveness of our dynamic opacity optimization both qualitatively and quantitatively, we present per-scene results and rendering samples for our method and the ablation model in \crefrange{tab:ablation_per_scene_synthetic}{tab:ablation_per_scene_tanks} and \crefrange{fig:ablation_bonsai}{fig:ablation_truck}, respectively. The ablation model is designed according to the experimental setup described in the main paper, where all parameters of prior layers, including opacity, are kept fixed during the training of enhancement layers. 

This comparison reveals that without dynamic opacity optimization, the model becomes significantly more redundant as additional splats are required to compensate for deficiencies in the lower layers. This results in larger model size and a decline in visual quality, particularly in areas that demand high-frequency detail. In contrast, our method dynamically refines the representation by adjusting the opacity of lower layers, ensuring that only essential splats contribute to the final output. This approach not only reduces model size but also enhances visual fidelity, especially in complex scenes. The results underscore the efficiency and scalability of our method across various datasets.

\begin{table*}[t]
    \centering
    \caption{Per-scene size and quality difference in percentage of {\name} on synthetic Blender dataset~\cite{mildenhall2021nerf} at different quality levels, compared to the Freeze method.}
    \label{tab:ablation_per_scene_synthetic}
    \resizebox{\linewidth}{!} {%
        \begin{tabular}{c|ccc|ccc|ccc}
            \hline 
            \multirow{2}{*}{ Scene } & \multicolumn{3}{c|}{$L_1$} & \multicolumn{3}{c|}{$L_2$} & \multicolumn{3}{c}{$L_3$} \\
            %
             & $\Delta$ SSIM $\uparrow$ & $\Delta$ LPIPS $\downarrow$ & $\Delta$ Size $\downarrow$ & $\Delta$ SSIM $\uparrow$ & $\Delta$ LPIPS $\downarrow$ & $\Delta$ Size $\downarrow$ & $\Delta$ SSIM $\uparrow$ & $\Delta$ LPIPS $\downarrow$ & $\Delta$ Size $\downarrow$ \\
            \hline 
            Lego & 0.21\% & -16.43\% & -113.78\% & 0.65\% & -40.72\% & -121.51\% & 1.19\% & 43.6\% & -232.51\% \\
            Hotdog & 0.25\% & -7.87\% & -128.50\% & 0.31\% & -9.39\% & -62.31\% & 0.18\% & 8.85\% & -87.35\% \\
            Ship & 0.32\% & -7.67\% & -68.57\% & 0.75\% & -10.25\% & -106.61\% & 3.88\% & 6.74\% & -146.29\% \\
            Materials & 0.35\% & -28.27\% & -96.70\% & 0.84\% & -39.19\% & -135.39\% & 1.30\% & 29.62\% & -233.48\% \\
            Ficus & 0.62\% & -59.32\% & -58.96\% & 1.59\% & -100.36\% & -96.93\% & 2.59\% & 108.65\% & -126.38\% \\
            Mic & 0.12\% & -34.28\% & -59.83\% & 0.18\% & -36.87\% & -68.31\% & 0.32\% & 42.37\% & -142.97\% \\
            Chair & 0.10\% & -23.60\% & -135.32\% & 0.28\% & -31.62\% & -110.90\% & 0.61\% & 28.68\% & -103.74\% \\
            Drums & 0.42\% & -13.39\% & -77.88\% & 1.06\% & -27.87\% & -96.12\% & 1.65\% & 32.42\% & -144.60\% \\
            \hline
        \end{tabular}
    }
    \vspace{0.2cm}
    \centering
    \caption{Per-scene size and quality difference in percentage of {\name} on Mip-NeRF360 dataset~\cite{barron2022mip} at different quality levels, compared to the Freeze method.}
    \label{tab:ablation_per_scene_360}
    \resizebox{\linewidth}{!} {%
        \begin{tabular}{c|ccc|ccc|ccc}
            \hline 
            \multirow{2}{*}{ Scene } & \multicolumn{3}{c|}{$L_1$} & \multicolumn{3}{c|}{$L_2$} & \multicolumn{3}{c}{$L_3$} \\
            %
             & $\Delta$ SSIM $\uparrow$ & $\Delta$ LPIPS $\downarrow$ & $\Delta$ Size $\downarrow$ & $\Delta$ SSIM $\uparrow$ & $\Delta$ LPIPS $\downarrow$ & $\Delta$ Size $\downarrow$ & $\Delta$ SSIM $\uparrow$ & $\Delta$ LPIPS $\downarrow$ & $\Delta$ Size $\downarrow$ \\
            \hline 
            Treehill & 0.93\% & -10.51\% & -72.26\% & 2.53\% & -21.85\% & -128.95\% & 17.78\% & -50.71\% & -124.24\% \\
            Room & 1.24\% & -17.90\% & -95.75\% & 2.94\% & -30.95\% & -180.00\% & 7.42\% & -50.87\% & -265.26\% \\
            Bonsai & 0.47\% & -3.20\% & -109.87\% & 0.86\% & -17.66\% & -247.38\% & 1.16\% & -27.92\% & -400.09\% \\
            Counter & 0.71\% & -3.57\% & -103.37\% & 1.52\% & -11.85\% & -208.83\% & 2.12\% & -24.54\% & -332.97\% \\
            Kitchen & 0.97\% & -7.92\% & -88.17\% & 1.98\% & -19.65\% & -181.91\% & 1.97\% & -16.39\% & -256.25\% \\
            Flowers & 1.64\% & -11.20\% & -79.76\% & 10.22\% & -20.21\% & -139.81\% & 26.00\% & -36.98\% & -154.45\% \\
            Garden & 0.58\% & -5.74\% & -87.92\% & 1.73\% & -13.61\% & -152.35\% & 12.22\% & -56.77\% & -195.91\% \\
            \hline
        \end{tabular}
    }
    \vspace{0.2cm}
    \centering
    \caption{Per-scene size and quality difference in percentage of {\name} on Deep Blending dataset~\cite{hedman2018deep} at different quality levels, compared to the Freeze method.}
    \label{tab:ablation_per_scene_db}
    \resizebox{\linewidth}{!} {%
        \begin{tabular}{c|ccc|ccc|ccc}
            \hline 
            \multirow{2}{*}{ Scene } & \multicolumn{3}{c|}{$L_1$} & \multicolumn{3}{c|}{$L_2$} & \multicolumn{3}{c}{$L_3$} \\
            %
             & $\Delta$ SSIM $\uparrow$ & $\Delta$ LPIPS $\downarrow$ & $\Delta$ Size $\downarrow$ & $\Delta$ SSIM $\uparrow$ & $\Delta$ LPIPS $\downarrow$ & $\Delta$ Size $\downarrow$ & $\Delta$ SSIM $\uparrow$ & $\Delta$ LPIPS $\downarrow$ & $\Delta$ Size $\downarrow$ \\
            \hline 
            Playroom & 0.51\% & 3.15\% & -126.67\% & 0.90\% & -10.87\% & -272.72\% & 1.29\% & -14.90\% & -378.33\% \\
            Drjohnson & -0.75\% & 7.00\% & -95.71\% & 3.04\% & -21.03\% & -168.14\% & 5.35\% & -27.28\% & -240.81\% \\
            \hline
        \end{tabular}
    }
    \vspace{0.2cm}
    \centering
    \caption{Per-scene size and quality difference in percentage of {\name} on Tank\&Temples dataset~\cite{knapitsch2017tanks} at different quality levels, compared to the Freeze method.}
    \label{tab:ablation_per_scene_tanks}
    \resizebox{\linewidth}{!} {%
        \begin{tabular}{c|ccc|ccc|ccc}
            \hline 
            \multirow{2}{*}{ Scene } & \multicolumn{3}{c|}{$L_1$} & \multicolumn{3}{c|}{$L_2$} & \multicolumn{3}{c}{$L_3$} \\
            %
             & $\Delta$ SSIM $\uparrow$ & $\Delta$ LPIPS $\downarrow$ & $\Delta$ Size $\downarrow$ & $\Delta$ SSIM $\uparrow$ & $\Delta$ LPIPS $\downarrow$ & $\Delta$ Size $\downarrow$ & $\Delta$ SSIM $\uparrow$ & $\Delta$ LPIPS $\downarrow$ & $\Delta$ Size $\downarrow$ \\
            \hline 
            Train & 1.61\% & -15.55\% & -74.58\% & 6.05\% & -26.54\% & -116.69\% & 11.00\% & -29.56\% & -145.92\% \\
            Truck & 1.75\% & -25.31\% & -75.40\% & 3.74\% & -68.70\% & -144.47\% & 5.90\% & -38.85\% & -164.20\% \\
            \hline
        \end{tabular}
    }
\end{table*}

